# Brain Tumor Detection and Classification Using a New Evolutionary Convolutional Neural Network

[1*]Amin Abdollahi Dehkordi, [1]Mina Hashemi, [2,3,4,5]Mehdi Neshat, [5]Seyedali Mirjalili, [6]Ali Safaa Sadiq

[1]Computer Engineering department, Najafabad Branch, Islamic Azad University, Najafabad, Iran, amin.abdollahi.dehkordi@gmail.com, Mina.Hashemi984@gmail.com

[2]Australian Centre for Precision Health, University of South Australia Cancer Research Institute, University of South Australia, Adelaide, SA, 5000, Australia, Mehdi.Neshat@unisa.edu.au, neshat.mehdi@gmail.com

[3] UniSA Allied Health and Human Performance, University of South Australia, Adelaide, SA, 5000, Australia

[4] South Australian Health and Medical Research Institute (SAHMRI), University of South Australia, Adelaide, SA, 5000, Australia

[5] Center for Artificial Intelligence Research and Optimization, Torrens University Australia, Brisbane 4006, Australia, ali.mirjalili@torrens.edu.au

[6]School of Engineering, Computing and Mathematical Sciences, University of Wolverhampton, Wulfurna Street, Wolverhampton, WV1 1LY, United Kingdom, Ali.Sadiq@wlv.ac.uk

**Abstract**

A definitive diagnosis of a brain tumour is essential for enhancing treatment success and patient survival. However, it is difficult to manually evaluate multiple magnetic resonance imaging (MRI) images generated in a clinic. Therefore, more precise computer-based tumour detection methods are required. In recent years, many efforts have investigated on classical machine learning methods to automate this process. Deep learning techniques have recently sparked interest as a means of diagnosing brain tumours more accurately and robustly. The goal of this study therefore is to employ brain MRI images to distinguish between healthy and unhealthy patients (including tumour tissues). As a result, an enhanced convolutional neural network is developed in this paper for accurate brain image classification. The enhanced convolutional neural network structure is composed of components for feature extraction and optimal classification. Nonlinear Lévy Chaotic Moth Flame Optimizer (NLCMFO) optimizes hyperparameters for training convolutional neural network layers. Using the BRATS 2015 data set and brain image datasets from Harvard Medical School, the proposed model is assessed and compared with various optimization techniques. The optimized CNN model outperforms other models from the literature by providing 97.4% accuracy, 96.0% sensitivity, 98.6% specificity, 98.4% precision, and 96.6% F1-score, (the mean of weighted harmonic value of CNN precision and recall).

## 1. Introduction

Every day, a great number of magnetic resonance (MR) pictures are generated as a result of technical advancement. These photos are used to monitor the physical state of humans. Their study by specialists, on the other hand, is a challenging task because it aids in the medical diagnosis and treatment of the patient's medical judgment. In other words, medical image analysis is an important stage in disease diagnosis and treatment. As a result, medical image analysis has turn out to be an exciting subject matter with a significant role in current clinical applications. Early brain tumour diagnosis is a difficult undertaking that serves as impetus for additional research.

Stroke lesions and cerebral tumours are difficult situations in medical imaging because their precise recognition has a significant impact on clinical diagnosis. A brain tumour is an abnormal cell that grows in or around the brain and causes changes in brain structure and behaviour. According to National Brain Tumor Foundation (NBTF) study studies, brain tumours are the leading cause of death worldwide and have more than tripled in the last three decades (El-Dahshan et al., 2014).

The human brain is distinguished by an organized complexity that makes analysis extremely difficult. Compared to a vast number of MR images, in addition, the analysis and viewing expertise are quite limited. Manual process in analyzing these images has various drawbacks, including the fact that it takes time. Furthermore, maintaining a high level of concentration throughout categorization is tiring, which contributes to an increase in the faulty detection accuracy. As a result, an automated system with high accuracy is needed to analyse MR images, and computer aided diagnosis (CAD) is a possible answer.

One of the remarkable attempt in utilizing AI in detecting brain tumour via MRI images is Deep learning (DL) architecture(Abd-ellah et al., 2019). DL is also employed in various CAD systems and medical imaging purposes, which shows the great potential of such techniques in automating diagnosis. On the other attempt, convolutional neural network (CNN) (Ghassemi et al., 2020) is one of the other methods that has maintained effective performance in classifying radiological images in recent decades. These types of techniques are deep learning algorithms that have been employed in a variety of applications, including pattern recognition (Litjens et al., 2017). These benefits prompted us to develop a CNN algorithm for brain tumour diagnosis using MRI images in this study.

Traditional machine learning methods solve problems by breaking them down into various parts, such as modelling the problem mathematically, gathering and clearing data, extracting and selecting features, training, optimization, and model evaluation. For large-scale problems, the lengthy process of modelling problems with these types of methods increases simulation time and computational expense. In contrast to these methods, DNN is a deep end-to-end learning method that may combine several processing steps into a single DNN to lower the cost of developing and extracting complicated features (Ertuğrul, 2020). To obtain accurate and competitive performance, DNN merely requires sufficient data and proper tweaking of its hyper-parameters.

The evolution of DNNs has advanced dramatically in the recent decade as data and processing capacity have increased. One of these networks' drawbacks was under-fitting in the absence of enough data to train them, which was remedied with the introduction of big data. Furthermore, as computational power has increased, it has been able to model and train networks with more complicated and high-dimensional structures. However, such networks also suffer from some limitations. DNN architecture design, in most circumstances, necessitates human interaction. DNN requires a substantial amount of labelled data for training, which renders them inapplicable in some instances where labelled data is difficult to obtain. This labelled data typically consumes large resources (for example, their lengthy labelling time) and, in addition, requires high labelling accuracy, which often comes at a cost. Furthermore, improved DNN performance is closely tied to researchers' repeated fine-tuning of their hyperparameters. These hyperparameters are critical to the accurate performance of DNNs. Fine-tuning them by researchers, on the other hand, is tough and time-consuming. As a result, optimising the values of these hyperparameters in these networks can be considered as a significant and effective challenge in how they perform.

Over the last decade, scientists and researchers have praised meta-heuristic (MH)(Dokeroglu et al., 2019) methodologies for their ability in solving real-world challenges. In general, the use of MH approaches to solve various issues has grown at an exponential rate. On the other hand, a heuristic algorithm is designed to solve a specific sort of problems. A meta-heuristic is a high-level method that can tackle a wider range of problems. Meta-heuristics use stochastic mechanisms to work with a set of randomly produced solutions to a given optimization problem. As a result, when using standard optimization methods is not feasible, they can find reasonably good solutions to a problem in an acceptable amount of time. Meta-heuristics have recently gained popularity for three reasons: 1) derivative independence, 2) reasonable computational time, and 3) high local optima avoidance. In other words, because such optimization algorithms treat optimization problems as black boxes, their applications are widespread in both research and industry.

The number of solutions generated randomly in the early stages of the algorithm divides meta-heuristic techniques into two groups: single-solution and multi-solution. In the first group, a solution is generated at random and get improved over the algorithm's iterations. Unlike single-solution approaches, multi-solution methods develop multiple solutions, and, with each iteration, the algorithm attempts to enhance the generated solutions.

Algorithms in the first group benefit from low computational cost as the objective function is only employed once in each iteration to assess the single solution. The convergence rate is also tend quick. Such methods, however, are prone to premature convergence. This refers to the trapping of a single solution in a locally optimal solution, as well as the algorithm's failure to resolve or achieve the global optima of the given optimization problem.

The use of a population of solutions in each iteration benefits multi-solution-based meta-heuristics. In comparison to single-solution meta-heuristics, premature convergence can be better handled due to information sharing and 'collaboration' amongst those solutions. Despite this advantage, they are typically more computationally costly because the objective function must be called for all solutions in each iteration. As a result, the convergence speed gets slowed down.

Despite multi-solution meta-heuristics have some intrinsic limitations, the literature demonstrates that such algorithms are now the primary method for solving optimization issues. This area of research is one of the most active in computational intelligence and has lately presented several methods. As a result of having such a potential opportunity in addressing a solution provided by a cost-effective optimization algorithm capable of equitably and efficiently finding the ideal solution, we were motivated to proposed an improved version of the recently-proposed Moth-Flame Optimization (MFO) (Mirjalili, 2015) algorithm.

The MFO algorithm simulates the transverse direction of moths, which is a typical motive in nature. Despite its virtues, it has been discovered that the exploratory and exploitative behaviours of this algorithm can be improved to address a broader range of issues. This spurred our efforts to incorporate nonlinear functions, the levy flight theorem, and chaotic values into the main stochastic components of this method, resulting in the introduction of a new version known as the Non-linear Lévy Chaotic Moth Flame Optimizer (NLCMFO) in this study.

A real-world optimization problem has been formulated to demonstrate the capacity of our proposed NLCMFO algorithm, in addition to evaluating it using a set of benchmark functions. CNN hyper-parameters have a substantial effect on network efficiency since they directly affect the training process. Choosing the appropriate hyper-parameters is critical in training CNN. For instance, when the learning rate is getting low, the network most likely going to miss out some of the key data details. In contrast, if the learning rate is excessively high, it could contribute in quick model convergence. As a result, CNN hyper-parameters must be optimized for optimum training and performance. The goal of this research therefore is to optimize the hyper-parameters of a CNN using the NLCMFO optimization algorithm, which is a new version of the Moth-Flame optimization (MFO) algorithm.

The remainder of this article is organized as follows: Section 2 demonstrates the related works. Section 3 presents the theoretical and mathematical principles that stand behind the proposed algorithm. Section 4 details the content and methodology of the proposed algorithm. Section 5 discusses the results of the conducted experiments in this study. Finally, Section 6 concludes the paper and presents the future steps.

## 2. Related work

Numerous experts in medical tomography and complex processing have obtained significant advancements in brain tumour identification over the recent years. There have been proposals for both fully automatic and semi-automatic solutions. The level of supervision involvement in the diagnosing procedures has mostly dictated the clinical acceptance of the used approach (Yanase & Triantaphyllou, 2019). The method of detecting the existence or absence of a tumour utilizing MRI image databases is known as detection. One of the most common issues in the healthcare profession is detection. The tumour detection method produces an MRI image that is designated as normal or abnormal (Rehman et al., 2020). To discover the best detection approach, a big standard database must be used as a benchmark to train the classifier and determine the optimum method for feature extraction and detection. As most lesions are small and have many changes in color intensity, shape, and texture, interpreting MR images by qualified physicians for the identification of a brain tumour surgical procedure is a time-consuming, and challenging undertaking. As a result, even well-trained neurologists or surgeons may find it difficult to make the right judgment. The other two reasons of misunderstanding are noisy visuals and a weary clinician. Quantitative image analytics are potential ways for coping with MRI scan defects. Deep learning architectures are the most prominent among machine learning methods. Although deep learning approaches for detecting brain tumours are a relatively new field of study, with few academic works on the subject, we covered in this part a brief synopsis of recently published studies on the diagnosis of brain tumours using machine learning (J. S. ; Paul et al., 2017).

Paul et al. (J. S. Paul et al., 2017) looked into the use of convolutional neural networks. They classified the photos using three classifiers: CNN, fully connected neural network (FCNN), and random forest. The CNN was configures with two convolutional layers (each with 64 filters of size $5 \times 5$), MaxPool layers, two fully connected layers with 800 neurons, and a SoftMax layer as the output layer.

Recently, an automated multi-modal diagnosis system was proposed in (Abd-Ellah et al., 2018) to detect and localize the different types of brain tumours using a deep CNN and an error-correcting output code combined with a support vector machine (ECOC-SVM). This hybrid model played the role of a feature classifier and extractor. In the second step of the proposed system from (Abd-Ellah et al., 2018), a deep CNN was used that includes five layers region-based (R-CNN) to localize the brain tumour within the abnormal MRIs. According to the results reported in (Abd-Ellah et al., 2018), the proposed system (used an AlexNet) achieved a high accuracy at 99.55%. It was also able to beat other non-deep learning models. Gao et al. (X. W. Gao et al., 2017) proposed 2D and 3D convolutional deep learning models with seven layers, and in each layer, the convolution and sub-sampling were applied to classify the CT brain image. The average classification precision was 87.6% in the best modelling. However, the CNN's hyper-parameters were not considered in these studies, which is essential to ensure maximum performance of the learning algorithm.

In another study, Rehmad et al. (Rehman et al., 2020) introduced a computer-aided tumour detection model consisting of three various CNN models (VGG16, GoogLeNet, and AlexNet) with an embedded transfer learning and data augmentation technique in order to classify the tumours of the brain. From (Rehman et al., 2020), the highest accuracy proposed by the VGG16 model was up to 98.69%. However, in most classification case studies, the performance of VGG19 was better than VGG16. Although deeper Convolutional Neural models can extract more complex nonlinear interactions and decode and enhance the model accuracy, most of them are faced with issues such as exploding or vanishing gradients. One of the most effective modifications is Residual Network (ResNet). ResNet is a popular CNN architecture introduced by He et al. in (He et al., 2016). One significant benefit of ResNet models is that it lets the training of intense CNNs using the shortcut connections in order to ignore one layer or more. One application of the ResNet for segmenting and classifying brain tumours based on MRI images can be seen in (Ismael et al., 2020); The applied ResNet with a set of augmentation methods such as zooming, shifting and rotating performed better than previous studies CNN (Abiwinanda et al., 2019; J. S. ; Paul et al., 2017), CapsNets (Afshar; et al., 2018, 2019), and KELM-CNN (Pashaei; et al., 2018).

Another instance of the application of modern convolutional deep learning models in detecting brain disorders is that Ehret et al. (Ehret et al., 2022) introduced an automated technique to detect and segment the multi-sequence brain MRI using a GoogLeNet architecture, and also the classification results represented that the GoogLeNet performance outweighed other machine learning methods. Later work by Amin et al. (Amin et al., 2020) used a stacked sparse auto encoder (SSAE) model with two fine-tuned layers hybridized with a high pass filter plus a median filter in recognizing brain tumours. Additionally, extensive experiments were used to adjust the hyper-parameters. However, the hyper-parameter optimizer in (Amin et al., 2020) was not fast and effective. In (Raj; et al., 2020), to develop an accurate classifier for detecting brain disorders and Alzheimer's disease, a hybrid deep learning model was formed consisting of Restricted Boltzmann Machine (RBM), optimal feature selection model (GLCM and GLRLM) cooperated with the Crow Search optimization approach. Based on the reported accuracy results in (Raj; et al., 2020), the proposed model proposed the highest accuracy, sensitivity and specificity compared with CNN, NN, and SVM. Nevertheless, the importance of the characteristics of the architecture and hyper-parameters were not assessed in (Raj; et al., 2020). Furthermore, Noreen et al. (Noreen; et al., 2020) presented the performance of the deep learning model can be improved using the pre-trained CNN models such as Inception-v3 and DensNet201 in the classification of brain tumour intensity. The findings in (Noreen; et al., 2020) represented that both models performed better than fine-tuned VGG16 (Rehman et al., 2020) and CNN-ELM (Pashaei; et al., 2018).

Recently, the application of meta-heuristic optimization algorithms has been substantially increased to improve the performance of medical deep learning models. An alternative approach was suggested by Hu et al. (Hu & Razmjooy, 2021) in diagnosing various types of brain tumours employing a deep belief network was combined with a meta-heuristic (seagull optimization algorithm) for improving the feature selection process. Another investigation of the meta-heuristics application in enhancing the performance of the brain tumours diagnosis was done by Aly et al. (Aly et al., 2019). They depicted a comparison of three optimisation algorithms: Ant colony (ACO) (Dorigo et al., 2006), Binary Particle Swarm Optimization (BPSO) (Khanesar; et al., 2007), and Artificial Bee colony (ABC) (Basturk & Karaboga, 2007) to develop the effectiveness of the group method of data handling (GMDH) (Amiri & Soleimani, 2021). The prediction results show that the ACO performance is better than the two other methods. In another recent research (Bezdan; et al., 2021), a hybrid system was proposed in classifying the glioma brain tumour, including a convolutional neural network that was modified by the elephant herding optimization (EHO) (Wang; et al., 2015) method. The EHO was applied to calibrate hyper-parameters of CNN, and also it resulted in a classification improvement rate. However, the computational budget was not reported in order to optimise the CNN's hyper-parameters. There are a considerable number of studies in the application of meta-heuristics to tune the hyper-parameters of the medical deep learning models. However, the majority of meta-heuristic optimizers have some control parameters that need to be initialized and adjusted during the optimization process. Initializing these control parameters is challenging, and we can see a high level of sensitivity in initializing these parameters with different values. Another issue associated with the control parameters of the meta-heuristic algorithm is that there is not a smart and effective strategy to update these parameters in the most of optimization methods. Meanwhile, providing an outstanding balance between the exploration and exploitation ability of the meta-heuristics is another critical challenge in this way.

**3. Preliminary considerations and concept**

This section describes the convolutional neural network, MFO, Levy Flight Theorem, and Chaotic Maps that will be employed to build our NLCMFO algorithm.

*3.1. Convolutional neural network concept*

The convolutional neural network (CNN), also known as ConvNet, has a highly advanced architecture and can generalise better than fully layered networks (Nebauer, 1998). Some of the reasons why CNN is more popular than other traditional models are as follows. First, the main reason for using CNN is the concept of weight sharing, which reduces the number of parameters required for training and thus improves generalisation (Arel et al., 2010). CNN can be trained quickly and easily with fewer parameters. Second, the classification stage is combined with the feature extraction stage (LeCun et al., 1998), which both employ a learning process. Third, implementing large networks using general artificial neural network (ANN) models is significantly more difficult than implementing CNN (Tivive & Bouzerdoum, 2005). Because of their superior performance, CNN networks are widely used in a variety of fields, including image classification (Donahue et al., 2014; Krizhevsky et al., 2017; Zeiler & Fergus, 2013), pattern recognition (Szegedy et al., 2013), face recognition (Timoshenko & Grishkin, 2013), speech recognition (Sainath et al., 2013), and many others. CNN is being considered as a model because of its ability to classify based on background information. A general CNN model is made up of three major layers: (a) the convolutional layer (CL), (b) the pooling layer (PL), and (c) the fully connected layer (FC). Each component's performance is depicted below.

- a. Convolutional layer (CL): The CNN's CL layer is in charge of extracting features from input images using a variety of convolution layers. The convolutional process is performed by these convolution layers at each offset of the input image. The CL has weights which must be optimised using gradient descent training that adjusts the CL's parameters. A nonlinear, rectified linear unit (ReLU) activation function is used to map the CL features into feature space. To normalise the gradients and activations via the network, a batch normalisation layer (BNL) is employed between both the CL and the ReLU.

b. Pooling layer (PL): After identification, the precise location of a feature becomes less important (Palsson et al., 2017). As a result, the pooling or sub-sampling layer comes after the convolution layer. The main advantage of using the pooling technique is that it reduces the number of teachable parameters while also introducing translation immutability (Fang et al., 2017). To perform the pooling operation, a window is chosen, and the input elements in that window are processed by a pooling function. Another output vector is generated by the pooling function. There are several pooling techniques, such as average collection and maximum collection, the most common of which is maximum pooling, which significantly reduces the size of the map. Because it does not participate in the forward flow, the error is not transmitted to the winning unit when calculating errors.

c. Fully connected layer (FC): A layer that is completely interconnected in standard models, it appears to be a fully connected network. The first stage's output (including convolution and max pooling) is fed to the fully connected layer, which calculates the product of the weight vector point and the input vector to obtain the final output. By evaluating the cost of the whole training data set and updating the variables after only one period, the descending gradient algorithm, generally described as batch mode learning or offline learning, decreases the cost function. Scrolling through the entire data set is covered in a course. These provide global minimums, but if the training data set is huge, the time taken to train the network will grow dramatically. A random descending gradient has been used to reduce the cost function in place of this method.

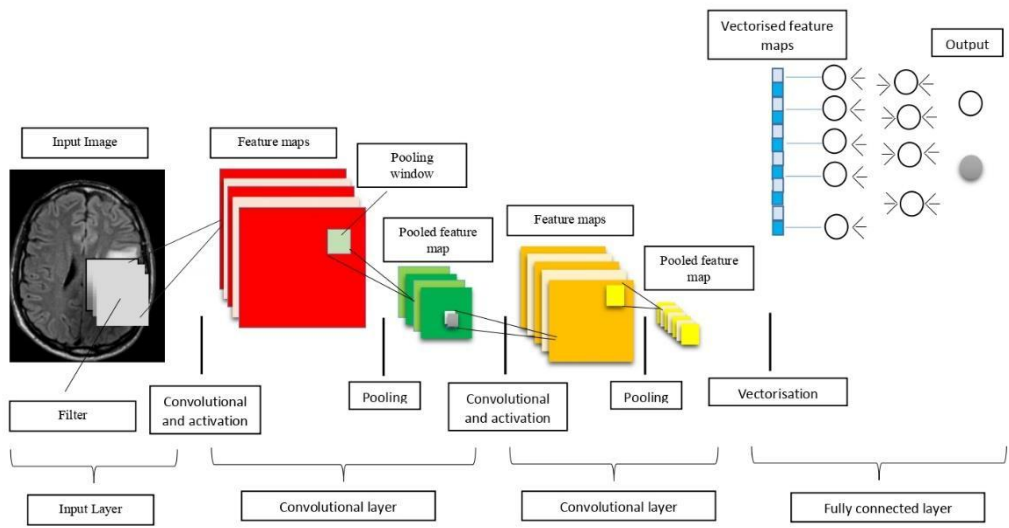

**Fig. 1.** A typical CNN Architecture(Lundervold & Lundervold, 2019)

*3.2. Moth Flame Optimization Algorithm*

Methods based on swarm intelligence are inspired by collective intelligence in nature (Amini et al., 2018). They imitate the behaviour of social insects, fish families, and bird flocks. These methods' primary advantages are their flexibility and strength (Blum & Li, 2008). MFO is a population-based method developed by Mirjalili in 2015 (Mirjalili, 2015) that mimics the transverse orientation for navigation technique used by moths at night. Moths fly at night, relying on moonlight to guide them while maintaining a constant angle of attack. In this algorithm, moth behaviour is simulated as a new optimization technique. Following figure illustrates a conceptual representation of this phenomenon.

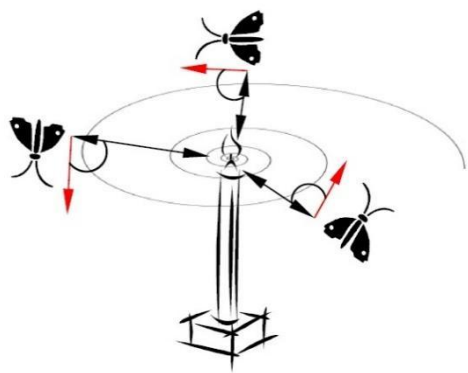

**Fig.2.** Spiral flying path (Mirjalili, 2015)

A population-based algorithm and a local search strategy are combined to produce an algorithm capable of both global exploration and local exploitation. MFO, like other meta-heuristic algorithms, is straightforward, adaptable, and simple to implement. As a result, recently MFO algorithm has been employed in solving range of different optimization problems (Jangir et al., 2016).

The MFO begins by randomly producing moths in the search space, then calculates each moth's cost (i.e., position) values and considers the optimal position with the flame. To update the better positions labelled by the flame, the moth positions are updated based on the spiral motion performance. The best new individual positions are updated, and the previous processes (moth positions are updated and new ones are created) are repeated until the end criteria are met.

*3.2.1. Generation of the initial moth population*

Mirjalili, as stated in (Mirjalili, 2015), assumed that any moth could fly in 1-D, 2-D, or 3-D space, or in too much space. The moth set can be expressed as follows:

$$M = [m_{1,1}\ m_{1,2}\ \cdots\ \cdots\ m_{1,d}\ m_{2,1}\ m_{2,2}\ \cdots\ \cdots\ m_{2,d}\ \vdots\ \vdots\ \vdots\ \vdots\ \vdots\ m_{n,1}\ m_{n,2}\ \cdots\ \cdots\ m_{n,d}]$$

In this equation, *n* denotes the number of moths and *d* signifies the number of dimensions in the solution space. In addition, the cost function values for all moths are stored in an array, as shown below:

$$OM = [OM_1\ OM_2\ \vdots\ OM_n] \qquad (2)$$

Flames are the last remaining components in the MFO. The flames in multidimensional space are depicted in the matrix below based on their cost function vector:

$$F = [F_{1,1}\ F_{1,2}\ \cdots\ \cdots\ F_{1,d}\ F_{2,1}\ F_{2,2}\ \cdots\ \cdots\ F_{2,d}\ \vdots\ \vdots\ \vdots\ \vdots\ \vdots\ F_{n,1}\ F_{n,2}\ \cdots\ \cdots\ F_{n,d}]$$
(3)

$$OF = [OF_1\ OF_2\ \vdots\ OF_n] \qquad (4)$$

It is worth noting that the feasible solutions are considered to be as moths and flames. The main differential here is in the method they been handled and updated every iteration. The moths here are used to represent real search agents that explore the search space. While on the other hand, flames are representing the best position for moths that have ever been discovered. In other words, flames can be compared to flags or pins thrown by moths while searching for better solutions in the search space. Accordingly, every moth seeks for a flag (flame) and updates it when a better solution is has been found. Hence, this method will mitigate any chance of losing the moth's best solution.

*3.2.2. Moth's positions updating*

To achieve global optimal convergence in optimization issues, MFO utilizes three distinct operations. The following are the definitions of these functions:

$$MFO = (I, P, T) \qquad (5)$$

Where *I* refers to the first random locations of the moths ($I : \phi \rightarrow \{M, OM\}$, *P* describes the movement of the moths in the search space ($P : M \rightarrow M$), and *T* to the end of the search process ($T : M \rightarrow true, false$)The *I* function, which is used to carry out a random distribution, is depicted in the following Equation.

$$M(i, j) = (ub(i) - lb(j)) * rand + lb(i) \quad (6)$$

Where *lb* and *ub* represent the variables' lower and upper bounds, respectively. Moths, as previously stated, fly across the search space in a transverse direction. When using a logarithmic spiral, there are three things to consider:

- The spiral's beginning point should be the moth.
- The flame's position should be the spiral's end point.
- The amplitude of the spiral oscillation should not exceed the search space.

Therefore, the logarithmic spiral for the MFO algorithm is as follows:

$$S(M_i, F_j) = D_i \cdot e^{bt} \cdot \cos\cos(2\pi t) + F_j \quad (7)$$

Where $D_i$ represents the distance between the *i-th* moth and the *j-th* flame (e.g., $D_i = |F_j - M_i|$), *b* denotes a constant value used to define the logarithmic helix shape, and *t* represents a random number between [1, -1]. The spiral flying pattern of moths is represented in Eq. (7). The next position of a moth is specified regarding a flame, as shown in this equation. The t value in the spiral equation specifies how close the moth's next position should be to the flame (t = -1 denotes the closest position to the flame, while t = denotes the maximum distance) (Mirjalili, 2015). The MFO's t-parameter is defined as follows.

$$t = (a - 1) * rand + 1 \quad (8)$$

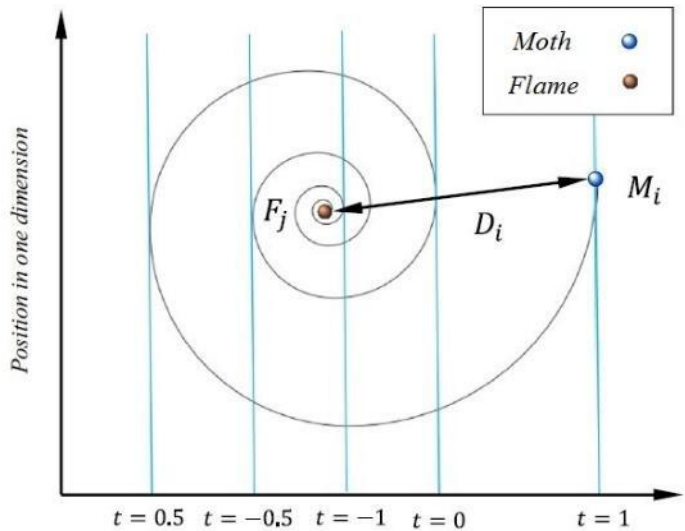

**Fig. 3.** Logarithmic spiral (Mirjalili, 2015).

Fig. 3 depicts the logarithmic spiral, the space around the flame, and the position assessing various t on the curve. The spiral motion of the moths near the flame in the search space ensures the balance between exploitation and exploration in MFO. Furthermore, optimal solutions are maintained each time the moths fly around the flames using the *OF* and *OM* matrices to avoid falling into the local optimal points (i.e., each moth flies in the vicinity of the nearest flame). Fig. 4 depicts a hypothetical model of a moth's position update around a flame. It should be noted that the vertical axis only represents one dimension (1 variable or parameter of a given problem), but the recommended technique can be used to adjust all of the variables of the issue.

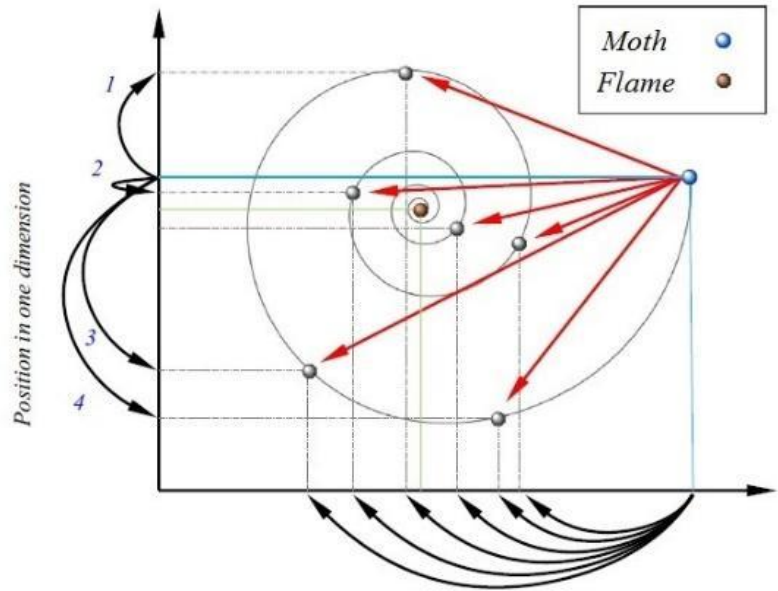

**Fig. 4.** some of the different locations a moth might take in relation to a flame utilizing the logarithmic spiral (Mirjalili, 2015).

*3.2.3 Update the number of flames*

This section focuses on the exploitation phase of the MFO (Updating the positions of moths in *n* distinct locations in the solution space, for instance, may diminish the likelihood of exploiting the highly viable solutions.). Therefore, as illustrated by the following Equation, minimizing the number of flames benefits in the solution of this issue:

$$flame\ no = round\left(N - l * \frac{N-l}{T}\right) \qquad (9)$$

Where $N$ represents the highest number of flames, while $l$ indicates the present iteration's count, and $T$ is the total number of iterations.

*3.2.4. Performance analysis of MFO*

Before discussing the methods used to improve the MFO algorithm's performance, it is necessary to examine its search pattern. Six benchmark functions are used to test the algorithm in this section. Fig.5 depicts how an MFO solves three unimodal functions with a single optimum and three multimodal functions with multiple optimizations. The figure demonstrates the structure of the solution landscape, the past observed spots of solutions, the obtained path/trajectory used by moths in finding solutions, the achieved average fitness by all solutions, and the convergence curves. The criteria mentioned above are explained further below. In this experiment, 80 different solutions were used for over 500 iterations.

- History of sampled points: This displays the moths' location history during optimization.
- The trajectory of one of the solutions: In each iteration, the path of the first moth shows the value of the first variable. Path curves show that in the early stages of optimization, the moths exhibit large and abrupt changes. A population-based meta-heuristic approach can ensure that this behaviour eventually converges to a point in the search space.
- Average fitness of all solutions: The average value of all moths in each iteration is shown here. If the curves show descending behaviour in all test functions, this behaviour proves that the algorithm improves the approximate optimal accuracy while performing the simulation.

Looking at the results in Fig. 5, we can observe that the search history for solutions is not as extensive occasionally. Despite the fact that a reasonable rational estimate of the global optimum point can be found in all test functions, the exploration potential should be expanded, especially for problems with many variables. This will be the first problem in this study that we will attempt to solve. Based on the trajectories, average fitness (objective values), and convergence curves in the test functions, the algorithm is abruptly transferred to exploration after approximately two-thirds of the iterations. This abrupt change may result in premature convergence. The transition from exploration to exploitation in meta-heuristic algorithms must be slow. This is the MFO's second drawback, which is addressed in the proposed method.

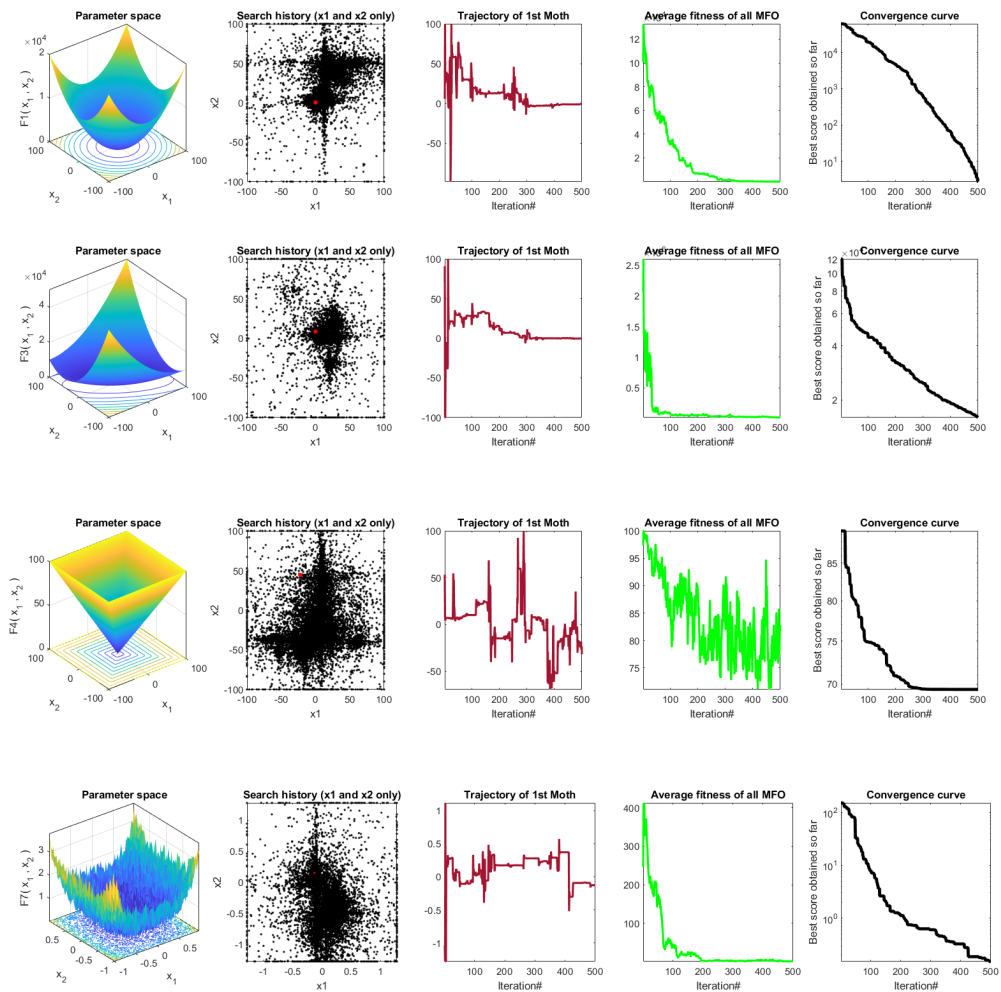

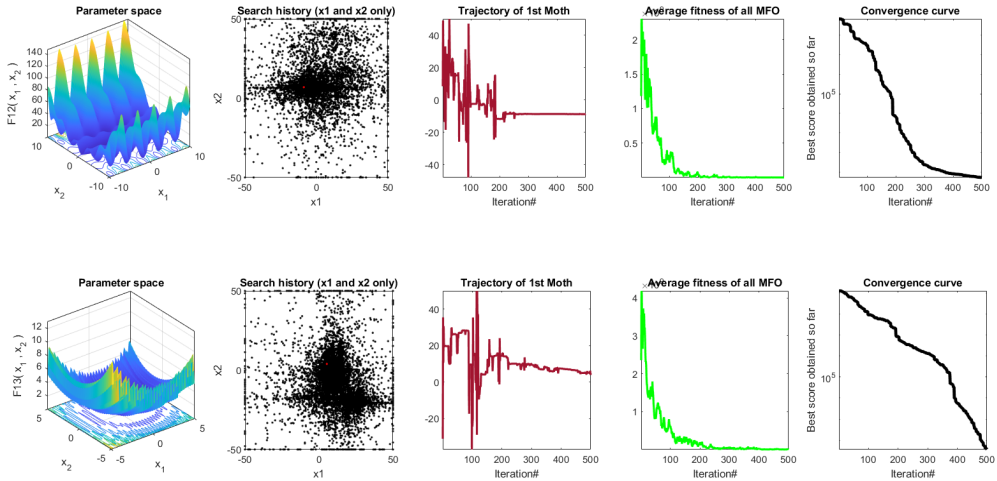

**Fig. 5.** Analysis of MFO search behaviour and performance when solving uni-modal and multi-modal test functions.

*3.3. Chaotic Maps*

The methods that use chaotic parameters instead of random parameters in random-based optimization techniques are referred to as "chaotic optimization algorithms (COA)". Due to the non-repetition and ergodicity of chaos, these algorithms can perform comprehensive searches at a higher velocity than stochastic searches that rely on probability (H. Gao et al., 2006). Chaos phenomenon is a definite or undefined process that exists in nonlinear, dynamic (non-periodic), non-convergent, and finite processes. To do this, one-dimensional, non-invertible maps are used to construct chaotic sets (Mitić et al., 2015). As the initial variable of these chaotic maps, any number in the range [1, 0] can be used. It should be noted, however, that the initial value has a significant impact on the oscillation pattern of chaotic maps (Zawbaa et al., 2016).

Recent advancements in theoretical and practical aspects of nonlinear dynamics, particularly chaos, have piqued the interest of researchers in a variety of domains. The use of chaos in optimization techniques is just one of those domains. During the last decade, chaotic maps have received considerable appeal throughout optimization experts due to their dynamic characteristics. This is because chaotic maps help the optimization technique improve its exploration of the search space. In other words, unlike a probability-based search, the exploration can be systematic instead of random (Wang et al., 2016). In addition, chaotic maps can help optimization algorithms avoid optimal local solutions and accelerate convergence (Ewees & Elaziz, 2020; Oliva et al., 2017; J. Xu et al., 2019; Yu et al., 2016). Table 1 lists the chaotic map equations. In Fig. 6, these equations are also illustrated to provide a more comprehensive understanding of their behaviour.

**Table 1** Chaos maps and their equations (Tavazoei & Haeri, 2007).

| Num. | Name | Equations |
|---|---|---|
| 1 | Logistic map | $X_{n+1} = aX_n(1 - X_n),\ a = 4$ |
| 2 | Tent map | $X_{n+1} = \{2X_n \quad X_n < 0.5\ \ 2(1 - X_n)\ X_n \geq 0.5$ |
| 3 | sinusoidal map | $X_{n+1} = aX_n^2 sin(\pi X_n)$ |
| 4 | circle map | $X_{n+1} = X_n + b - \left(\frac{a}{2\pi}\right) \sin sin\left(2\pi X_n\right) mod(1), a = 0.5\ and\ b$ |

| Num. | Name | Equations |
|---|---|---|
| 5 | Gauss map | $X_{n+1} = \{0 \quad X_n = 0 \quad \frac{1}{X_n \bmod(1)} \quad otherwise$ |
| 6 | Chebyshev map | $X_{n+1} = cos(a.cos^{-1}X_n)$ |
| 7 | Singer map | $X_{n+1} = (7.86X_n - 23.31X_n^2 + 28.75X_n^3 - 13.302875X_n^4), a =$ |
| 8 | Sine | $X_{n+1} = \frac{a}{4}\sin\sin(\pi X_n), a = 4$ |
| 9 | Iterative | $X_{n+1} = \sin\sin(\frac{a\pi}{X_n}), a = 0.7$ |

Fig. 2 visualized these nine chaotic maps. Each sub-figure in Fig. 6 is a chaotic map of the relevant equation from Table 1. Fig. 6 demonstrates the behaviour patterns of logistic, tent, sinusoidal, circle, Gauss, Chebyshev, Singer, Sine, and iterative maps using 2D diagrams. The chaotic map values are represented on the y-axis, while the number of iterations is represented on the x-axis. As evidenced by the chaotic maps' movement, there are more possibilities for the areas covered and the number of times they move. As a result, employing the certain map enhances the exploration behaviour of an optimization techniques. It is also worth mentioning that these maps were considered since they exhibit diverse behaviours when generating chaotic values and have been demonstrated to be effective in numerous experiments. (Gandomi et al., 2013; Wang et al., 2013).

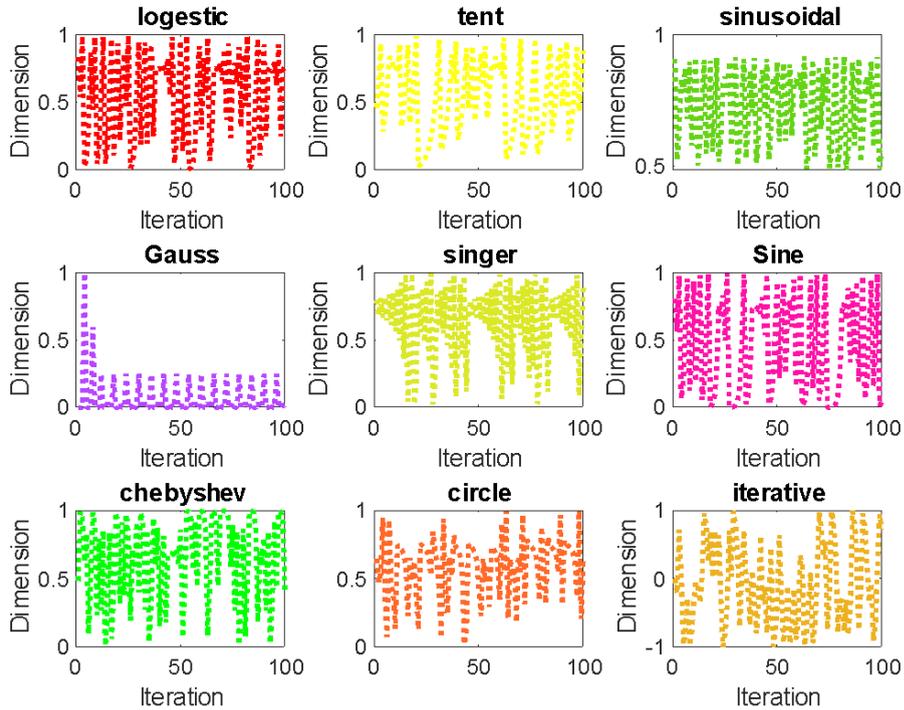

**Fig. 6**. Visual perception of chaotic maps.

*3.4. Lévy Flight Theory*

Flight performance developed Lévy flight theory, which is a stochastic step method. This function will control the step size depending on the probability distribution function generated by the Lévy distribution (power-sequence law) (Faramarzi et al., 2020):

$$L(x_i) \approx |x_i|^{1-\alpha} \qquad (9)$$

The power level (power-law exponent) is denoted by $1 < \alpha \leq 2$ when the flight length is denoted by $x_i$. Mantegna, on the other hand, has proposed a fast and accurate algorithm for producing a more stable Lévy process. This method yields an arbitrary index distribution (α) value between 0.3 and 1.99. Given these benefits, our proposed method generates random numbers based on the Lévy distribution as follows.

$$Levy(\alpha) = 0.05 \times \frac{x}{|y|^{\frac{1}{\alpha}}} \qquad (10)$$

Where x and y are characterized by Eqs. (9, 10), the two normal distribution variables are defined by the deviation of the $\sigma_x$ and $\sigma_y$ criteria, as follows:

$$x = Normal(0, \sigma_x^2) \qquad (11)$$

$$y = Normal(0, \sigma_y^2) \qquad (12)$$

While the standard deviations are described in Eqs. (11, 12), the variables $\sigma_x$ and $\sigma_y$ are defined as follows:

$$\sigma_x = \left[\frac{\Gamma(1+\alpha)sin(\frac{\pi\alpha}{2})}{\Gamma(\frac{(1+\alpha)}{2})\alpha 2^{\frac{(\alpha-1)}{2}}}\right]^{1/\alpha}, \sigma_y = 1, \alpha = 1.5 \qquad (13)$$

Where α is produced between 0.3 and 1.99. However, in this study, its value is adjust to 1.5. Fig. 7 is illustrated the Levy flight's movement pattern.

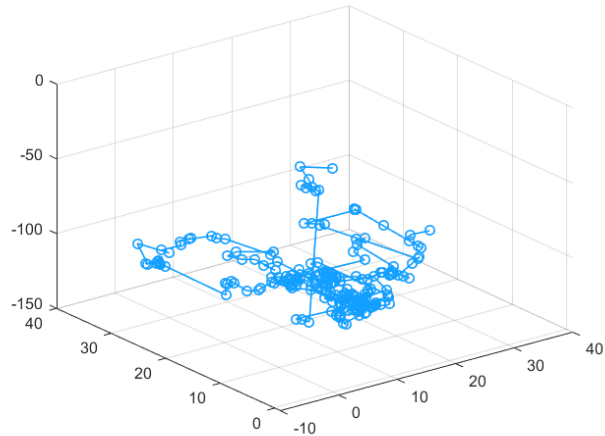

**Fig.** 7. A modeled α = 1.5 Levy flight (LF) movement pattern

Lévy flight, as depicted in the Fig. 7, is generally connected with small steps and, on rare occasions, long jumps. This property can be used to promote search efficiency in the optimization field, due to more efficiency performance in comparison to uniform random search (Deb, 1997; Yang & Deb, 2013). The movements from chaotic maps, which are plainly visible in Fig. 6, on the other hand, can explore and investigate remote portions of the search space but cannot search as correctly and profoundly as the Lévy approach. Thus, this study attempts to present an efficient optimization method called nonlinear levy chaotic Moth Flame Optimization (NLCMFO) that benefits from the unique characteristics of the Lévy strategy as well as features of chaotic maps. It is worth mentioning that a search space can be investigated and exploited both globally and locally by combining these methods and taking use of the special characteristics of each strategy.

## 4. Proposed Method

The proposed NLCMFO is presented in this section. The primary purpose of the NLCMFO is to increase the performance of the standard MFO in two segments. The initial stage is to integrate MFO with Levy flight theory and chaotic maps, followed by employing the nonlinear weight coefficient parameter as a control variable between both the algorithm's exploration and exploitation processes.

*4.1. NLCMFO Algorithm*

As stated in section 3.4, MFO lacks adequate exploration and exploitation processes. Therefore, the primary goal of the NLCMFO algorithm is to address this obvious flaw in the MFO. In MFO, Eq. (7) simulates the spiral flight path of moths. The next position of a moth is defined by this equation based on the flame. In the spiral motion equation, the parameter t determines how close the moth's next position should be to the flame (t = -1 indicates the closest position to the flame, while t = 1 indicates the furthest distance). As a result, an excess ellipse can be assumed to exist in all directions around the flame, and the next position of the moth in this space. Spiral motion is an important component of the MFO because it determines how the moths' positions around the flames are updated. It is also important mentioning that each moth will be able to fly around a flame not only in the space between them with the help of spiral equation, which provides this motion's behaviour. Hence, utilizing this type of motions, MFO could maintain both exploration and exploitation of the problem's searching space. Exploration occurs in MFO when the next position is outside the moth and flame space, and exploitation occurs when the next position is within the moth and flame space. To emphasize exploitation even further, *t* is a random number in [*a*, 1]. Where *a* decreases linearly from -1 to -2 during the iterations.

The MFO's t-parameter is critical in enhancing the algorithm's ability to locally explore for potential solutions to optimization issues and in boosting the algorithm's exploitation and exploration processes. However, as discussed in Section 3.4, this parameter has failed to play its role in creating appropriate exploration and exploitation processes for the MFO. The proposed method attempts to strengthen the role of this parameter in improving the exploration process with the least amount of change by substituting the chaotic values produced by the selected chaotic maps for the random values in Eq. (8). Then, Eq. (8) is updated as follows.

$$t = |(a - 1) * Cm + 1| \qquad (14)$$

The parameter *Cm* in Eq. (14) represents the chaotic values generated by the selected chaotic maps in the proposed method.

The question that may arise here is whether it is sufficient for the moths to move towards the flame to update the position in Eq. (7). As a result of this process, the MFO quickly becomes stuck in local optimizations. To avoid this, each moth in Eq. (7) must update its position using only one of the flames. The flames are classified based on the fitting values obtained after each iteration and after updating the flame list. The moths then update their position according to the corresponding flames. The moth always updates its position based on the best flame on the list, while the last moth updates its position based on the worst flame on the list. Each moth is assigned a specific flame to prevent it from becoming trapped in the local optimum. Since moths can only fly towards the flame and not around it, if all of them are attracted to the same flame, they will all converge at some point in the search space. Forcing them to move around different flames, on the other hand, leads to more exploration of the search space and a lower chance of being trapped in the optimal local locations. According to the above, using chaotic values and modifying Eq. (8) can improve the range of motion of the moths and their movement around different flames, leading to improved exploratory behaviour in the proposed method. To increase the chances of finding better solutions, the best solutions obtained thus far are referred to as flames. Therefore, the *F* matrix in Eq. (3) always contains the *n* best recently obtained solutions.

To generate chaotic values, the proposed method employs two distinct chaotic maps. Therefore, in the NLCMFO, if the values of the flame matrix do not change in several consecutive iterations due to the possibility of trapping NLCMFO in the local optimal, the chaotic values generated by the second chaotic map will be used in the next iteration. Moreover, this method ensures that the search space is explored in the best places obtained thus far for the following reasons:

- Moths in the upper ranges adjust their position based on the best solutions obtained thus far.
- The flame matrix shifts based on the best solutions in each iteration, and the moths must update their positions in response to the new flames. Thus, the use of various chaotic values updates the position of the moths around different flames, a mechanism that causes the moths to move abruptly in the search space and enhances exploration.

The t-parameter in the MFO affects not only the exploration but also how the algorithm can be exploited. By defining a new parameter, such as Levy flight theory, the proposed method attempts to increase the role of this parameter only in improving the exploration process and exploitation phase done with Levy flight theory. It should be noted that the exploration and exploitation capabilities in population-based meta-heuristic algorithms are at opposite ends of the spectrum, which means that improving one weakens the other. By defining a new parameter, such as Levy flight theory, the proposed method attempts to increase the role of this parameter only in improving the exploration process and exploitation phase done with Levy flight theory.

Therefore, the NLCMFO, in addition to improving exploration capability, also improves exploitability by employing Levy flight theory. Another important parameter in the proposed method is a nonlinear weight coefficient parameter, which is in charge of ensuring an appropriate equilibrium between the exploration and exploitation processes in NLCMFO. The following is how this parameter is expressed:

$$w = 4 * \exp\exp\left(-\left(6 * \frac{Iter}{Max_{Iter}}\right)^2\right) \qquad (15)$$

Where the value of *w* decreases nonlinearly between [4, 0]. Thus, Eq. (7) has been updated as follows.

$$S(M_i, F_j) = w * \vec{R}_L \otimes D_i \cdot e^{bt} \cdot \cos\cos(2\pi t) + \vec{R}_L \otimes F_j \qquad (16)$$

$\vec{R}_L$ is the matrix generated by Levy's flight theory which is denoted by the input multiplication sign in Eq. (16).

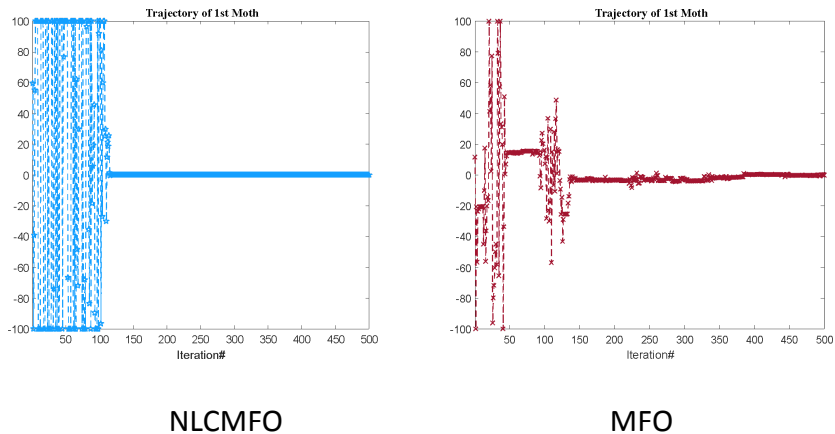

NLCMFO                              MFO

**Fig. 8.** Moth's position update in NLCMFO versus MFO

The exploration and exploitation phases in MFO are not as extensive and efficient as in NLCMFO, as seen in these figures. Therefore, it has been demonstrated that chaotic maps and the levy flight theorem can be used to modify these phases in MFO. Moths in the NLCMFO search the solution space both globally (exploration phase) and locally (exploitation phase) to exploit an interesting region and obtain the global optimum. The NLCMFOs' flowchart is demonstrated in Fig. 9.

*4.2. The NLCMFO_CNN architecture*

A convolutional neural network (CNN) is a deep learning paradigm that boosts the functionalities of an artificial neural network (ANN) by extending layers to its structure. These networks provide superior feature extraction, pattern recognition, and the ability to classify raw data without prior preprocessing. The CNN architecture is divided in two sections: feature extraction blocks and classification blocks. In the initial phase, multiple layers, particularly a convolutional (CL) and a max-pooling layer (MPL), are used to provide a feature extraction process. To classify the collected features from the first block, CNN employs a fully connected layer (FCL) and a classification layer.

Weights and biases in the convolutional and fully connected layers must be adjusted using a descending gradient training algorithm. Hyper-parameters that strongly affect the performance of the CNN model must also be included in the training algorithm. The parameters also include the training algorithm, momentum, learning rate, number of epochs, validation frequency, and L2Regularization. The primary goal of this research is to optimize these parameters for CNN training phase in order to achieve the best results possible using the NLCMFO algorithm. Despite the fact that these hyper-parameters are crucial for enhanced performance, assessing each one is computationally costly. As a result, in this architecture, NLCMFO is developed to optimize hyper-parameters for training CNN layers. The proposed network flow diagram optimized NLCMFO_CNN is illustrated in Fig. 10.

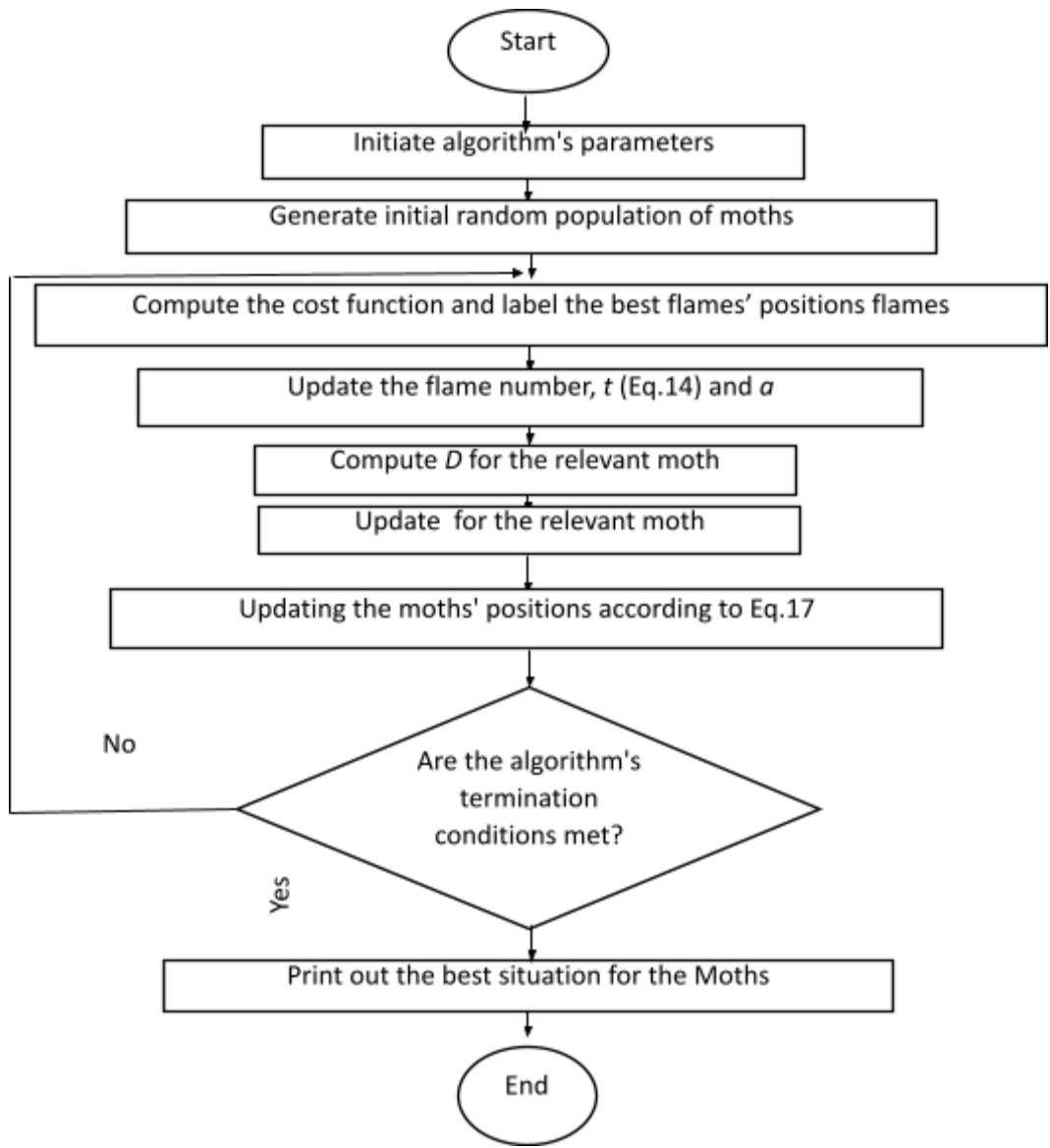

**Fig. 9.** The NLCMFO's flowchart.

The CNN training process, as detailed in Section 3, comprises modifying the settings of kernels and neurons concealed in a fully connected classification layer. In general, CNNs adjust the convolutional layer and fully connected layer parameters using stochastic descending gradient (SGD) training. By updating network weights in the opposite way, SGD minimizes the cost function. One drawback of using SGD is that it has many hyper-parameters that affect network performance.

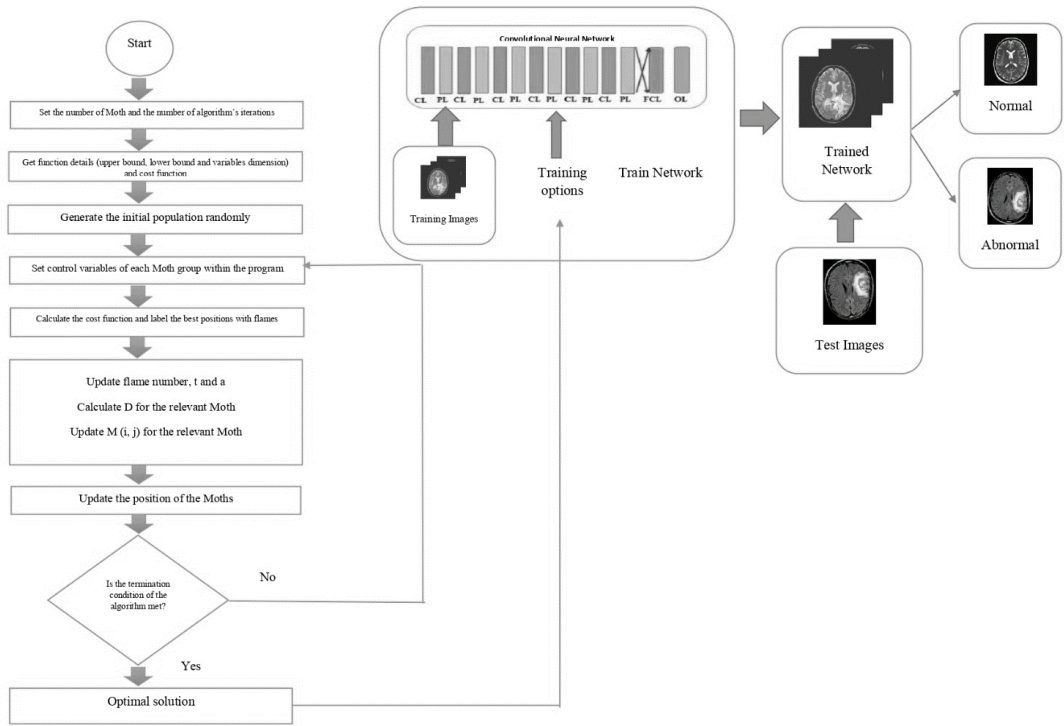

**Fig. 10.** Flow diagram of NLCMFO-CNN for brain tumor classification.

The following section explains how to use the NLCMFO algorithm to optimize these hyper-parameters for optimal network performance. Table 2 contains information on the proposed CNN layers.

**Table 2** The proposed NLCMFO-CNN architecture

| Layer | Input | Type | | | |
|---|---|---|---|---|---|
| | | 128×128×3 | | | |
| 1 | ConvL | Type | Filter size | No .of filters | stride |
| | | ConvL + BN+ ReLu | 7 × 7 | 128 | 1 × 1 |
| | Max Pooling Layer | | 2 × 2 | - | 2 × 2 |
| 2 | ConvL | ConvL + BN+ ReLu | 7 × 7 | 64 | 1 × 1 |
| | Max Pooling Layer | | 2 × 2 | - | 2 × 2 |
| 3 | ConvL | ConvL + BN+ ReLu | 7 × 7 | 64 | 1 × 1 |
| | Max Pooling Layer | | 2 × 2 | - | 2 × 2 |
| 4 | ConvL | ConvL + BN+ ReLu | 7 × 7 | 32 | 1 × 1 |
| | Max Pooling Layer | | 2 × 2 | - | 2 × 2 |
| 5 | ConvL | ConvL + BN+ ReLu | 7 × 7 | 16 | 1 × 1 |
| | Max Pooling Layer | | 2 × 2 | - | 2 × 2 |
| 6 | ConvL | ConvL + BN+ ReLu | 7 × 7 | 8 | 1 × 1 |
| | Max Pooling Layer | | 2 × 2 | - | 2 × 2 |
| 7 | FC | Type | | | |
| | | Output size =2 | | | |
| 8 | Output | Classification layer Soft-max | | | |

*4.4.1. Optimization of hyper-parameters using NLCMFO*

NLCMFO was used to optimize hyper-parameters in this study. Hyper-parameters heavily influence CNN's accuracy and convergence. The selection of network hyper-parameters is critical and is determined by the application for which CNN is used. The most common training hyper-parameters in convolutional neural networks are learning rate, epochs, momentum, and regulatory coefficient. The learning rate and momentum are controlled by the speed of the descending gradient algorithm and the effect of updating previous weights on updating current weights, respectively. The number of epochs determines how many times the learning algorithm's network parameters are updated based on the training dataset. The regularization coefficient keeps the network from becoming over-fit. Therefore, in order to address all of these settings, optimization of these hyper-parameters is required to help the network achieve the most accurate results. Algorithm (1) provides the NLCMFO algorithm for optimizing the parameters in SGD training.

---

**Algorithm 1.** Pseudo-code of Optimised NLCMFO-CNN

---

**Require:** Batch Size: $m$; number of search agents: $n$; dimensions: $d$; number of optimization iterations: $it$;
    hyper parameters: $K1, K2, K3, K4$; hyper parameter evaluation function: $L_D$

**Initialize:** Random NLCMFO Population $Xi\ (i = 1, 2, \ldots n)$.
    Initialize $a, t$

**Procedure:** Sample a batch of training data.
    **for** number of optimization iterations, $it$
      **for** each search agent
        Calculate fitness function
        Choose and tag the best position by flames
        Update flame number, $t, a$
        Calculate $D$ for the corresponding moth
        Calculate $M(i, j)$ for the corresponding moth
      **end for**
      (K1, K2, K3, K4) = hyper parameter evaluation function($L_D$)
    **end**
    $w\ \&\ b \leftarrow Sgdm\ (K1, K2, K3, K4)$
**end Procedure**

---

The convolutional neural network's $L_D$ evaluation function is defined as follows.

$$L_D = (1 - accuracy) * 100 \quad (17)$$

This function's task is to calculate the error rate after training the network hyper-parameters with the optimal values obtained by the NLCMFO algorithm in each iteration. In other words, this evaluation function serves as the NLCMFO algorithm's cost function.

$$accuracy = sum\left(\left(Y_P == Y_T\right) \div \left(Total\ number\ of\ Y_T\right)\right) \quad (18)$$

The $Y_P$ parameter in Eq. 17 represents the network output, and the $Y_T$ parameter indicates the desired network output. The NLCMFO algorithm calculates the evaluation function and the network error rate in each iteration of the proposed algorithm based on the proposed hyper-parameters. This function is computed for the proposed algorithm's entire population. Finally, the main hyper-parameters are chosen based on those with the lowest error rate in network training.

*5. Experimental Results*

This section discusses the research method for this study followed by the results. The research method is quantitative, as are other methods in meta-heuristic algorithms. The proposed algorithm is tested on a variety of experimental functions, which are discussed further in the following section. Because of the random nature of NLCMFO, this algorithm is run 30 times, and its performance is reported using descriptive statistical measures.

NLCMFO was compared to a number of well-known optimization algorithms in the literature to validate the results. Moreover, to verify that the results are not produced randomly, statistical standards and relevance assessments are applied in such comparisons. Scalability is also considered to maximize NLCMFO's performance. Finally, the performance analysis will be demonstrated and discussed in order to solve an optimal convolutional neural network for the automatic classification of brain images. It should be noted that the data sets employed in this study are a collection of standard mathematical functions that have been frequently considered in the literature to assess the efficiency of optimization techniques(Y. Xu et al., 2019).

*5.1. Benchmark functions details*

Numerous tests have been carried out in this section to prove the validity of the proposed strategy in relation to the conceptual points stated throughout the previous sections. Reviewing the literature, a well-known set of specialized benchmark functions was chosen, which contains 23 uni-modal, multi-modal, fixed-state dimensions multi-modal functions, and six composite functions that are often employed to test the efficiency of optimization methods (Rashedi et al., 2009; Yao et al., 1999).

The first category contains unimodal functions (F1-F7), which provide a single optimal solution and are used to assess the algorithm's capability to exploit them. The second group contains multi-modal functions (F8–F13) that have multiple optimal solutions. Local optimal solutions put the algorithm's exploration capability in these functions to the test. Meta heuristic algorithms, on the other hand, must be either able to explore the space globally while avoiding becoming trapped in the local optimal in order to identify the global optimal. Other categories that are similar to multi-modal functions but have lower and fixed dimensions are fixed-dimensional multi-modal functions (F14–F23). The F24-F29 functions include the composite functions published at CEC 2014 (Liang et al., 2014). These functions are generated by integrating displacement, rotation, and expansion along with the highest complex mathematical optimization problems available in the literature. Appendix A contains the mathematical formulation and attributes of these functions.

*5.2. Experimental setup*

The proposed NCLMFO is compared to a number of existing algorithms in the literature, including PSO(Eberhart & Kennedy, 1995), Whale Optimization Algorithm (WOA) (Mirjalili & Lewis, 2016), Salp Swarm Algorithm (SSA) (Mirjalili et al., 2017), Multi-Verse Optimization (MVO) algorithm (Mirjalili et al., 2016), MFO (Mirjalili, 2015), and Grey Wolf Optimizer (GWO) (Mirjalili et al., 2014). These algorithms' parameter settings are identical to the main algorithm proposed in the main article for that algorithm (see Table 3).

**Table 3** The algorithms' parameter adjustments

| Algorithm | Parameter | Value |
|---|---|---|
| PSO | Topology fully connected Inertia factor | 0.5 |
| | $c_1$ | 2 |
| | $c_2$ | 2 |
| MVO | Wormhole Existence Probability at its Maximum | 1 |
| | Wormhole Existence Probability at its Minimum | 0.2 |
| MFO | *a (Convergence constant)* | $[-1 \ -2]$ |
| SSA | $c_1$ (The exploration and exploitation phases' balancing variable) | [1 0] |
| GWO | *a (Convergence constant)* | [2 0] |
| WOA | *a (Convergence constant)* | [2 0] |

Matlab R2020b, which is installed on a PC running 64-bit Windows 8.1 and 6 GB of RAM, is employed to perform the NLCMFO and other standard methods. Furthermore, the initial population and iterations of all optimization techniques are limited to 30 and 500, respectively.

*5.3. NLCMFO performance assessment*

The benchmark functions were run, and the performance of each algorithm was measured, in order to determine the NLCMFO's performance and evaluate it to other advanced optimization techniques. As stated in section 4.3, the suggested approach enhances exploration capabilities by utilizing two chaotic maps, attaining the greatest performance with any F1-13 set fitting function. Therefore, the NLCMFO employs two chaotic maps, Sine and Chebyshev. A value of 0.7 is chosen as the initial condition for these two chaotic maps. Furthermore, these maps have been adjusted in the interval [1, 0] to be applied in the MFO's Eq. (8) instead of random values.

*5.4. Qualitative results*

This section presents and analyzes the results of a qualitative analysis of the NLCMFO for solving single-state and multi-state functions. During the implementation of the NLCMFO, qualitative assessment evaluates positional shifts and also significant changes in the fit function conceptually. The four well-known indicators are depicted in Fig. 11: search history, the trajectory of one of the solutions, the average fitness of all solutions, and the convergence curves. One of the solutions' trajectories analyzes how the first-factor variable changes during the optimization procedure. The average fitness graph illustrates how the overall population fit improves as optimization progresses. The convergence graph displays that NLCMFO has found a promising fit with rather uniform convergence at the completion of the iteration procedure.

Although the NLCMFO does not have as extensive a search history as MFO in some functions in Fig. 11, it can be seen that it can achieve the global optimum with a far higher efficiency than MFO. According to the trajectory diagram, the NLCMFO solutions saw rapid shifts in the early phases before collapsing in the latter phases. This behavior virtually ensures that the NLCMFO will finally settle on a place and exploit the region.

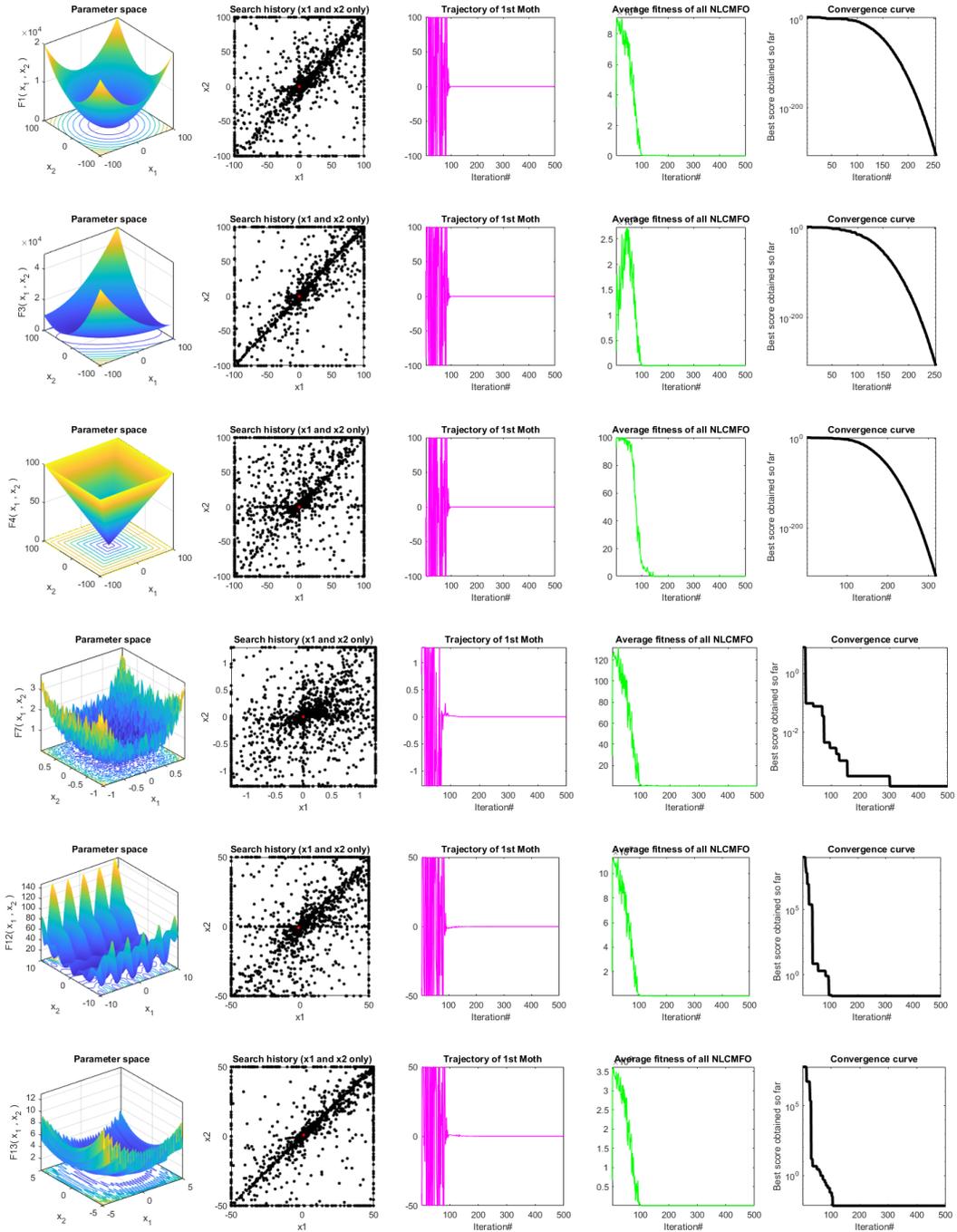

**Fig.11.** Qualitative results for the benchmark functions F1, F3, F4, F7, F12 and F13.

The trajectory seen in Fig. 11 demonstrates the algorithm's early exploratory activity as a result of unexpected motions. The NLCMFO's fast convergence and highly efficient near-optimal global exploration are ensured by the sudden intensity in the early iterations and the partial intensity in future iterations, and the NLCMFO is encouraged to shift from exploration to exploitation patterns (Kharrufa et al., 2018). This analysis also demonstrates the proposed NLCMFO behaviour for exploration. By reviewing the mean of the fit curve provided in Fig. 11, the improvement of NLCMFO fit conditions throughout iterations revealed the favorable influence of Levy flight theory, chaotic map values, and the nonlinear weight coefficient parameter. While the average NLCMFO fit in some conditions shows a constant downward curve and decreases, indicating an improvement in the population's overall output.

*5.4. Scalability analysis*

This section examines and discusses the scalability of the proposed NLCMFO. The NLCMFO performance of several benchmark functions is assessed employing scalability analysis. However, it also demonstrates how a meta-heuristic algorithm may keep its exploration strength even while dealing with large-scale challenges. Because these algorithms are random, the results of a single run of the method are erroneous. Therefore, each method is performed 30 times with varied variables to eliminate unpredictability. Furthermore, statistical data has been acquired. Standard deviations have been recorded in conjunction to the mean, and all related results are summarized in Tables 4 through 6. The demonstrated mean (Ave) and standard deviation (Std) of the results of all optimization techniques were gathered and examined for each function across 30 runs of 500 iterations.

*5.4.1. Exploitation assessment*

The results in Table 4 clearly show that NLCMFO comes in first when solving F1–F7. The proposed algorithm has a distinct advantage in solving F1–F7 functions among the implemented versions. The results demonstrate that NLCMFO surpasses other techniques in nearly all experimental functions, particularly the MFO. This capacity is derived from the suggested exploitation phase's tiny motions described by the Levy flight characteristic and a nonlinear weight coefficient variable that determines the NLCMFO's exploitation and exploration stages.

*5.4.2. Exploration assessment*

The results in Table 4 demonstrate that NLCMFO is still efficient in multi-modal functions. Compared to other algorithms, the proposed method's average for F8–F11 was the smallest. Furthermore, NLCMFO results in F12 and F13 are acceptable and significantly smaller than PSO, indicating that NLCMFO also can outperform advanced algorithms as well as the proposed algorithm's ability to avoid displaying the optimal local solution.

**Table 4** benchmark functions Results (F1 -F13)

| Func. | | WOA | SSA | GWO | PSO | MVO | MFO | **NLCMFO** |
|---|---|---|---|---|---|---|---|---|
| F1 | Ave | 3.12E-73 | 5.61E-07 | 8.70E-28 | 1.85E-04 | 1.34E+00 | 2.34E+03 | **0.00E+00** |
| | Std | 1.65E-72 | 2.00E-06 | 9.05E-28 | 1.98E-04 | 3.67E-01 | 4.30E+03 | **0.00E+00** |
| F2 | Ave | 2.31E-51 | 1.63E+00 | 9.48E-17 | 4.82E-02 | 8.20E+00 | 2.80E+01 | **0.00E+00** |
| | Std | 8.10E-51 | 1.27E+00 | 8.03E-17 | 6.38E-02 | 2.73E+01 | 1.92E+01 | **0.00E+00** |
| F3 | Ave | 4.43E+04 | 1.49E+03 | 1.21E-05 | 8.54E+01 | 2.23E+02 | 1.81E+04 | **0.00E+00** |
| | Std | 1.92E+04 | 1.02E+03 | 3.22E-05 | 3.10E+01 | 9.14E+01 | 9.89E+03 | **0.00E+00** |
| F4 | Ave | 5.08E+01 | 1.17E+01 | 5.34E-07 | 1.62E+00 | 1.79E+00 | 6.91E+01 | **0.00E+00** |
| | Std | 2.64E+01 | 3.25E+00 | 5.34E-07 | 2.56E-01 | 6.51E-01 | 7.60E+00 | **0.00E+00** |
| F5 | Ave | 2.80E+01 | 2.58E+02 | 2.70E+01 | 8.97E+01 | 6.66E+02 | 1.50E+04 | **6.99E+00** |
| | Std | 4.01E-01 | 4.04E+02 | 8.21E-01 | 9.03E+01 | 8.74E+02 | 3.11E+04 | **3.41E+00** |

| Func. | | WOA | SSA | GWO | PSO | MVO | MFO | **NLCMFO** |
|---|---|---|---|---|---|---|---|---|
| F6 | Ave | 3.76E-01 | **1.59E-07** | 8.03E-01 | 1.57E-04 | 1.27E+00 | 9.94E+02 | 3.36E-03 |
| | Std | 2.21E-01 | **1.09E-07** | 3.98E-01 | 1.47E-04 | 2.94E-01 | 3.02E+03 | 5.97E-03 |
| F7 | Ave | 3.59E-03 | 1.62E-01 | 2.29E-03 | 1.75E-01 | 3.34E-02 | 3.52E+00 | **8.03E-05** |
| | Std | 3.88E-03 | 5.77E-02 | 1.02E-03 | 6.02E-02 | 1.53E-02 | 6.83E+00 | **8.74E-05** |
| F8 | Ave | **-1.07E+04** | -7.25E+03 | -5.94E+03 | -4.87E+03 | -7.72E+03 | -8.59E+03 | -3.93E+03 |
| | Std | **1.79E+03** | 6.26E+02 | 1.04E+03 | 1.27E+03 | 5.33E+02 | 9.61E+02 | 4.82E+02 |
| F9 | Ave | 1.89E-15 | 4.77E+01 | 2.64E+00 | 5.53E+01 | 1.23E+02 | 1.61E+02 | **0.00E+00** |
| | Std | 1.04E-14 | 1.44E+01 | 4.10E+00 | 1.14E+01 | 2.56E+01 | 3.00E+01 | **0.00E+00** |
| F10 | Ave | 3.26E-15 | 2.64E+00 | 1.00E-13 | 2.95E-01 | 2.55E+00 | 1.68E+01 | **8.88E-16** |
| | Std | 2.35E-15 | 1.07E+00 | 1.64E-14 | 5.14E-01 | 3.26E+00 | 5.22E+00 | **0.00E+00** |
| F11 | Ave | 2.57E-02 | 1.43E-02 | 4.74E-03 | 5.11E-03 | 8.60E-01 | 1.33E+01 | **0.00E+00** |
| | Std | 9.54E-02 | 1.12E-02 | 8.39E-03 | 7.84E-03 | 9.54E-02 | 3.10E+01 | **0.00E+00** |
| F12 | Ave | 2.45E-02 | 6.95E+00 | 4.43E-02 | **1.38E-02** | 2.08E+00 | 1.30E+01 | 1.73E-02 |
| | Std | 2.37E-02 | 3.89E+00 | 2.17E-02 | **4.50E-02** | 1.02E+00 | 1.80E+01 | 1.33E-02 |
| F13 | Ave | 4.62E-01 | 1.68E+01 | 5.75E-01 | **5.56E-03** | 2.25E-01 | 1.37E+07 | 2.70E-02 |
| | Std | 1.97E-01 | 1.33E+01 | 2.44E-01 | **6.15E-03** | 1.59E-01 | 7.49E+07 | 1.49E-02 |

According to the results, incorporating chaotic values within the MFO not only expands the algorithm's exploration stage but also improves its efficiency.

**Table 5** Benchmark function results (F14–F23)

| Func. | | WOA | SSA | GWO | PSO | MVO | MFO | NLCMFO |
|---|---|---|---|---|---|---|---|---|
| F14 | Ave | 2.86E+00 | 1.20E+00 | 4.59E+00 | 2.77E+00 | 9.98E-01 | 2.25E+00 | 1.00E+00 |
| | Std | 3.27E+00 | 6.05E-01 | 4.18E+00 | 2.46E+00 | 4.36E-11 | 1.78E+00 | 1.25E-02 |
| F15 | Ave | 7.10E-04 | 8.86E-04 | 4.48E-03 | 8.80E-04 | 3.45E-03 | 2.29E-03 | 6.67E-04 |
| | Std | 5.52E-04 | 2.48E-04 | 8.08E-03 | 2.73E-04 | 6.75E-03 | 4.93E-03 | 1.70E-03 |
| F16 | Ave | -1.03E+00 | -1.03E+00 | -1.03E+00 | -1.03E+00 | -1.03E+00 | -1.03E+00 | -1.03E+00 |
| | Std | 9.87E-10 | 1.42E-14 | 2.01E-08 | 6.25E-16 | 5.34E-07 | 6.78E-16 | 1.34E-04 |
| F17 | Ave | 3.98E-01 | 3.98E-01 | 3.98E-01 | 3.98E-01 | 3.98E-01 | 3.98E-01 | 4.00E-01 |
| | Std | 1.69E-05 | 0.00E+00 | 2.29E-06 | 0.00E+00 | 3.62E-07 | 0.00E+00 | 9.40E-03 |
| F18 | Ave | 3.00E+00 | 3.00E+00 | 3.00E+00 | 3.00E+00 | 3.00E+00 | 3.00E+00 | 3.00E+00 |
| | Std | 2.82E-04 | 2.96E-13 | 2.75E-05 | 1.36E-15 | 3.58E-06 | 2.90E-15 | 1.37E-03 |
| F19 | Ave | -3.86E+00 | -3.86E+00 | -3.86E+00 | -3.86E+00 | -3.86E+00 | -3.86E+00 | -3.86E+00 |
| | Std | 1.34E-02 | 1.83E-10 | 2.45E-03 | 2.64E-15 | 3.55E-06 | 1.44E-03 | 9.20E-03 |
| F20 | Ave | -3.20E+00 | -3.25E+00 | -3.27E+00 | -3.28E+00 | -3.28E+00 | -3.23E+00 | -3.21E+00 |
| | Std | 1.75E-01 | 6.62E-02 | 7.37E-02 | 5.83E-02 | 5.95E-02 | 6.11E-02 | 9.05E-02 |
| F21 | Ave | -8.94E+00 | -7.56E+00 | -8.47E+00 | -7.47E+00 | -7.21E+00 | -5.97E+00 | -1.00E+01 |
| | Std | 2.18E+00 | 3.12E+00 | 2.65E+00 | 3.02E+00 | 3.11E+00 | 3.17E+00 | 4.58E-01 |
| F22 | Ave | -7.28E+00 | -8.23E+00 | -1.04E+01 | -7.80E+00 | -8.52E+00 | -7.28E+00 | -1.04E+01 |
| | Std | 3.04E+00 | 3.18E+00 | 1.18E-03 | 3.31E+00 | 3.03E+00 | 3.47E+00 | 6.28E-02 |
| F23 | Ave | -7.05E+00 | -8.63E+00 | -9.81E+00 | -9.33E+00 | -8.23E+00 | -7.88E+00 | -9.86E+00 |
| | Std | 3.46E+00 | 3.27E+00 | 2.24E+00 | 2.76E+00 | 3.18E+00 | 3.58E+00 | 1.71E+00 |

The results in Table 5 show that the proposed algorithm performed very well against the standard functions F14–F23. However, compared to other methods, the proposed algorithm produces the best overall global results for F14–F23. Due to the results reported, it is possible to determine that meta-heuristic methods may effectively handle optimization issues if there is an acceptable amount of exploration and exploitation, as well as a logical equilibrium throughout the two stages of exploration and exploitation. The NLCMFO improves exploitation and exploration by leveraging the strengths of Levy flight theory and chaotic values. Furthermore, the proposed algorithm employs a nonlinear weight coefficient parameter to attempt to balance the two stages of exploration and exploitation. The NLCMFO strategy has been validated due to the results achieved with the suggested method when compared to other methods.

Dealing with composite mathematical functions is extremely difficult for meta-heuristic optimization algorithms due to the inherent difficulty. In order to solve these challenges, the exploration and exploitation phases in these algorithms must be well balanced.

**Table 6** Composite function results (CEC2014)

| Func. | | GWO | PSO | MVO | SSA | WOA | MFO | **NLCMFO** |
|---|---|---|---|---|---|---|---|---|
| F24 | Ave | 2.60E+03 | 2.63E+03 | 2.64E+03 | 2.64E+03 | 2.61E+03 | 2.68E+03 | **2.50E+03** |
| | Std | 3.41E-02 | 8.49E+00 | 5.72E+00 | 7.34E+00 | 6.64E+00 | 2.83E+01 | **0.00E+00** |
| F25 | Ave | 2.71E+03 | 2.72E+03 | 2.71E+03 | 2.72E+03 | 2.71E+03 | 2.72E+03 | **2.60E+03** |
| | Std | 5.07E+00 | 4.63E+00 | 2.26E+00 | 5.51E+00 | 1.90E+01 | 1.16E+01 | **0.00E+00** |
| F26 | Ave | 2.75E+03 | 2.78E+03 | 2.74E+03 | 2.70E+03 | 2.74E+03 | 2.70E+03 | **2.70E+03** |
| | Std | 6.09E+01 | 4.11E+01 | 6.08E+01 | 1.24E-01 | 6.50E+01 | 1.24E+00 | **0.00E+00** |
| F27 | Ave | 3.46E+03 | 3.49E+03 | 3.41E+03 | 3.70E+03 | 3.94E+03 | 3.65E+03 | **2.90E+03** |
| | Std | 1.09E+02 | 3.08E+02 | 1.88E+02 | 1.09E+02 | 2.97E+02 | 1.76E+02 | **0.00E+00** |
| F28 | Ave | 4.13E+03 | 6.73E+03 | 4.15E+03 | 4.28E+03 | 5.68E+03 | 3.95E+03 | **3.00E+03** |
| | Std | 4.04E+02 | 7.93E+02 | 3.59E+02 | 5.10E+02 | 5.66E+02 | 2.30E+02 | **0.00E+00** |
| F29 | Ave | 4.76E+06 | 4.44E+03 | 4.36E+05 | 6.79E+06 | 1.84E+07 | 3.14E+06 | **3.10E+03** |
| | Std | 7.92E+06 | 1.25E+03 | 2.27E+06 | 1.07E+07 | 2.09E+07 | 4.09E+06 | **0.00E+00** |

The results shown in Table 6 demonstrate the proposed algorithm's complete superiority over other methods in the face of these functions. Furthermore, the results demonstrate the presence of an appropriate NLCMFO equilibrium between both the exploration and exploitation processes, as well as the effectiveness of achieving this equilibrium utilizing the nonlinear weight coefficient variable in the NLCMFO.

*5.4.3. NLCMFO's computational complexity*

Computational complexity is an essential measure for determining an algorithm's runtime, which can be characterized depending on the structure and execution of the algorithm. The computational complexity of the MFO is determined by the number of moths, variables, the maximum number of iterations, and the flame sorting mechanism per iteration (Shehab et al., 2019). Because the Quicksort algorithm is used, the complexity of sorting calculations is $O(nlogn)$ at best and $O(n^2)$ at worst. Taking the $P$ function into account, the overall computational complexity is defined as follows:

$$O(MFO) = O(t(O(Quicksort) + O(Position\ update))) \qquad (17)$$

$$O(MFO) = O\left(t\left(n^2 + n \times d\right)\right) = O\left(tn^2 + tnd\right) \qquad (18)$$

Where $n$ is the number of moths, $t$ is the maximum number of iterations and $d$ is the number of variables.

On the other hand, the timing of the implementation of meta-heuristic methods to find a desirable proper solution is an important consideration that must be taken into account when evaluating the proposed algorithm against such a criterion.

**Table 7** Compare the average runtime (seconds) in more than 30 run.

| Func. | | WOA | SSA | GWO | PSO | MVO | MFO | NLCMFO |
|---|---|---|---|---|---|---|---|---|
| F1 | Ave | 2.01E+00 | 2.13E+00 | 3.73E+00 | 1.44E+00 | 4.10E+00 | 2.36E+00 | 2.66E+00 |

|  | Std | 5.52E-02 | 6.22E-02 | 6.98E-02 | 5.25E-02 | 2.24E-01 | 3.81E-02 | 9.95E-02 |
|---|---|---|---|---|---|---|---|---|
| F2 | Ave | 2.04E+00 | 2.18E+00 | 3.74E+00 | 1.38E+00 | 2.43E+00 | 2.34E+00 | 2.59E+00 |
|  | Std | 8.93E-02 | 7.87E-02 | 8.16E-02 | 6.31E-02 | 1.21E-01 | 6.89E-02 | 1.16E-01 |
| F3 | Ave | 1.42E+01 | 1.42E+01 | 1.56E+01 | 1.34E+01 | 1.59E+01 | 1.45E+01 | 1.44E+01 |
|  | Std | 2.88E-01 | 1.94E-01 | 2.01E-01 | 1.97E-01 | 1.68E-01 | 7.93E-01 | 1.63E-01 |
| F4 | Ave | 2.19E+00 | 2.16E+00 | 3.67E+00 | 1.39E+00 | 4.04E+00 | 2.25E+00 | 2.57E+00 |
|  | Std | 5.58E-02 | 6.80E-02 | 4.95E-02 | 1.52E-01 | 1.09E-01 | 8.64E-02 | 1.13E-01 |
| F5 | Ave | 2.12E+00 | 2.20E+00 | 3.79E+00 | 1.48E+00 | 4.21E+00 | 2.44E+00 | 2.69E+00 |
|  | Std | 7.26E-02 | 5.79E-02 | 5.36E-02 | 4.16E-02 | 1.03E-01 | 6.56E-02 | 9.85E-02 |
| F6 | Ave | 2.05E+00 | 2.10E+00 | 3.68E+00 | 1.44E+00 | 4.11E+00 | 2.33E+00 | 2.62E+00 |
|  | Std | 8.62E-02 | 6.11E-02 | 4.51E-02 | 4.37E-02 | 1.18E-01 | 7.65E-02 | 1.08E-01 |
| F7 | Ave | 3.68E+00 | 3.66E+00 | 5.18E+00 | 2.90E+00 | 5.61E+00 | 3.84E+00 | 4.26E+00 |
|  | Std | 6.46E-02 | 1.23E-01 | 5.05E-02 | 6.62E-02 | 9.87E-02 | 6.44E-02 | 1.13E-01 |
| F8 | Ave | 2.47E+00 | 2.57E+00 | 4.16E+00 | 1.94E+00 | 2.93E+00 | 2.76E+00 | 3.05E+00 |
|  | Std | 6.13E-02 | 9.10E-02 | 6.45E-02 | 5.65E-02 | 8.48E-02 | 7.40E-02 | 8.29E-02 |
| F9 | Ave | 2.29E+00 | 2.41E+00 | 3.84E+00 | 1.72E+00 | 4.53E+00 | 2.68E+00 | 2.83E+00 |
|  | Std | 7.62E-02 | 6.31E-02 | 3.72E-02 | 6.07E-02 | 1.02E-01 | 9.10E-02 | 1.02E-01 |
| F10 | Ave | 2.23E+00 | 2.47E+00 | 4.01E+00 | 1.81E+00 | 4.57E+00 | 2.71E+00 | 2.84E+00 |
|  | Std | 1.08E-01 | 7.82E-02 | 1.15E-01 | 5.10E-02 | 1.17E-01 | 9.30E-02 | 1.47E-01 |
| F11 | Ave | 2.53E+00 | 2.62E+00 | 4.15E+00 | 1.99E+00 | 4.74E+00 | 2.92E+00 | 3.12E+00 |
|  | Std | 8.70E-02 | 7.55E-02 | 8.65E-02 | 6.14E-02 | 1.11E-01 | 7.07E-02 | 8.85E-02 |
| F12 | Ave | 5.70E+00 | 5.77E+00 | 7.48E+00 | 5.14E+00 | 7.98E+00 | 6.13E+00 | 6.40E+00 |
|  | Std | 7.97E-02 | 5.43E-02 | 3.63E-02 | 3.19E-02 | 1.42E-01 | 5.93E-02 | 8.54E-02 |
| F13 | Ave | 5.84E+00 | 5.84E+00 | 7.48E+00 | 5.27E+00 | 7.93E+00 | 6.17E+00 | 6.27E+00 |
|  | Std | 7.12E-02 | 7.00E-02 | 8.32E-02 | 2.53E-02 | 1.04E-01 | 1.16E-01 | 8.80E-02 |

Table 7 summarizes all optimization execution times for F1-F13 optimization issues with 1000 dimensions to assess the performance of the NLCMFO and confirm the impact of employing Levy flight theory and chaotic values on its execution time. When examining Table 7, it is evident that NLCMFO has a good execution time. In comparison to MFOs, however, the employment of Levy flight and chaotic variables had no effect on the time necessary to identify the optimal overall optimization. While compared to other optimizers, the NLCMFO algorithm can acquire a reasonable and fair execution time when solving uni-modal and multi-modal functions, as well as the search space for big solutions, by examining the gathered results. This success can be attributed to the unique structure of meta-heuristic algorithms when it comes to dealing with optimization problems. In general, these observations imply that using Levy flight values, chaotic values, and a nonlinear weight coefficient parameter improves the effectiveness of the proposed approach.

*5.4.4. Convergence rate of the proposed method*

To evaluate the NLCMFO's convergence performance, the best value achieved thus far is determined and considered to represent the intended value. This value is computed and provided for each iteration of all the test functions used, and it was obtained by employing each of the benchmark approaches. It was fully documented, as were the NLCMFO's average results, which were logged and rigorously analysed across 30 runs and 500 iterations for each approach. In the convergence debate, as illustrated in Fig. 12, the NLCMFO has a fair convergence rate in most cases when compared to other approaches.

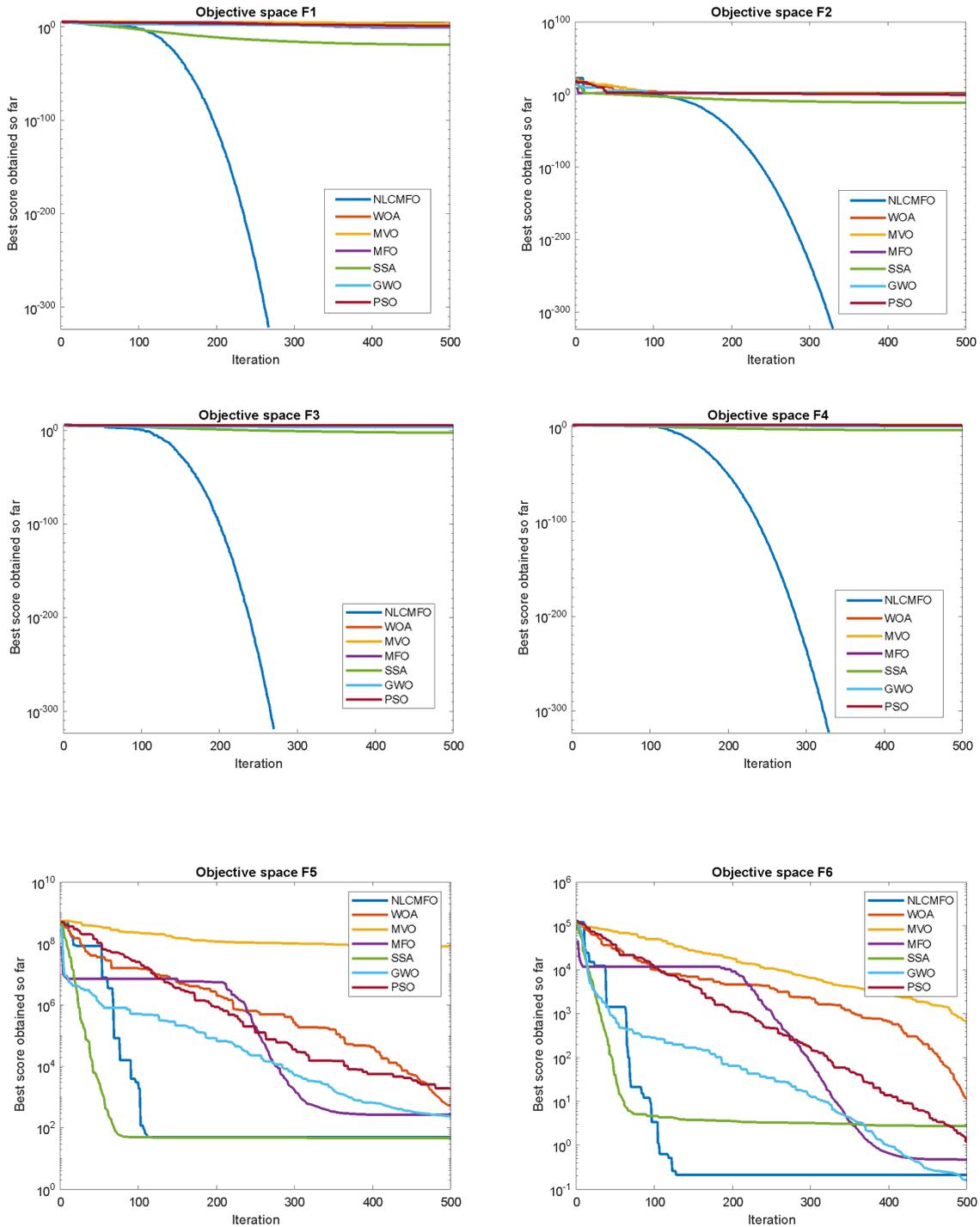

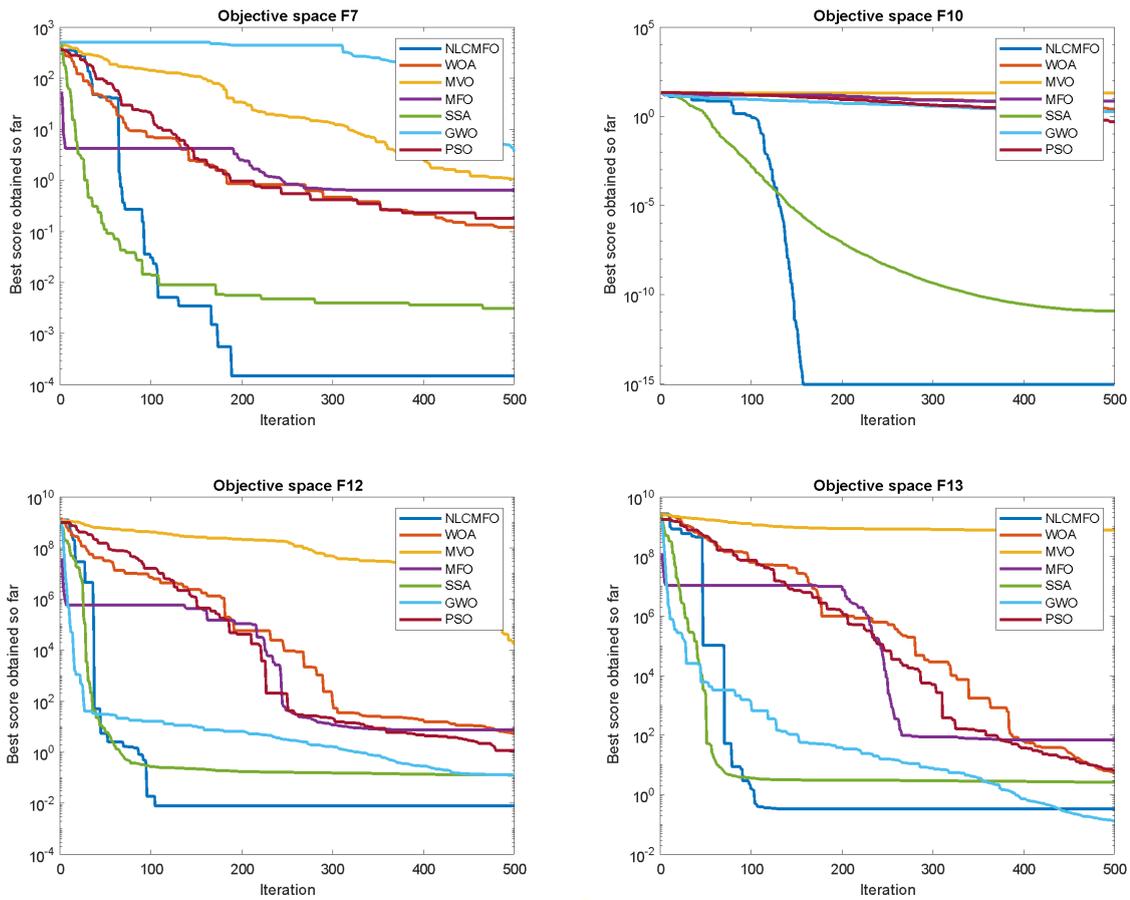

**Fig. 12.** NCLMFO versus other approaches convergence analysis

The reported results demonstrate and validate the dependability of Levy flight and chaotic values, which enhances the performance and convergence velocity of the optimization method proposed by the presented procedure. The NCLMFO, on the other hand, obtained the desired values in several functions that were relatively near to the values acquired by the other algorithms. The NCLMFO not only surpassed the other methods in the convergence curve, but it also obtained the optimal value in the fewest number of iterations and displayed a substantially faster convergence velocity than the other algorithms.

According to the results obtained by the proposed method to improve the performance of the proposed method using Levy flight theory, chaotic values and nonlinear weight coefficient parameters were proven, and the proposed method adequately addressed the disadvantages of the MFO. According to the results presented in this section, the nonlinear weight coefficient parameter defined in the proposed method played a critical role in achieving a proper balance between the proposed algorithm's exploration and exploitation phases. Furthermore, the application of Levy flight theory and chaotic values has considerably boosted the overall performance of the proposed method's exploration and exploitation.

*6. Performance analysis of NCLMFO-CNN*

Experiments were carried out to classify normal and abnormal images with tumors in order to compare the performance of CNN based on NLCMFO for the diagnosis of brain tumors.

## 6.1. NCLMFO-CNN Implementation Specifics

The network was built with MATLAB 2020b and runs on Windows 10 Pro with 16 GB of Nvidia GPU RAM. BRATS 2015 (*BRATS: The Virtual Skeleton Database Project.*, 2015) was used to test the proposed architecture for normal and abnormal MRI images. Other images have been gathered from the Harvard Medical School website (http://www.med.harvard.edu/aanlib/home.html).

### 6.1.1. Training phase

A convolutional neural network with six convolutional layers is employed to classify MRI brain images. 70% of the available data is consumed for training phase, while the other 30% is used for network testing. The architecture proposed divides images into two categories: normal and abnormal (with tumors). Randomly, the data set is separated into training and testing sets. Using data augmentation, all image sizes are resized to $128\times128\times3$ and converted to color images. During the training phase, the parameters are adjusted with the SGD training algorithm, with the hyper-parameters optimized using NLCMFO. The NLCMFO_CNN optimized variables are listed below. Fig. 13 shows the sample training examples.

- Training algorithm: SGD
- Momentum: 0.5849
- Initial learning rate: 0.0337
- Maximum epoch: 10
- L2Regularization: 1.2959e-04

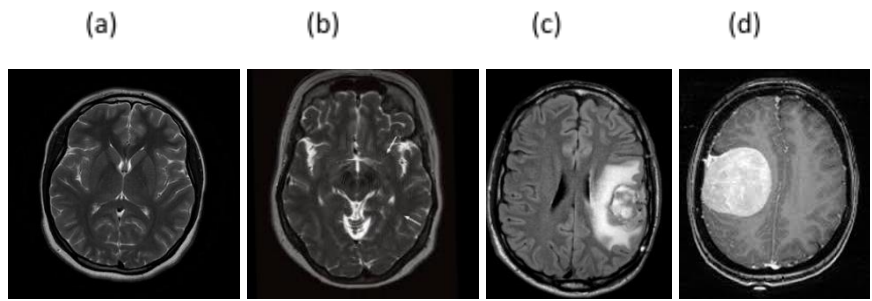

**Fig. 13.** Examples of training images (a), (b) normal images, (c), (d) abnormal images (included tumors)

### 6.1.2. Testing phase

Using the data augmentation process and the proposed training network, the image size was changed to 128×128×3 at this point. The test images are then fed into a CNN-trained model, with all CL and FCL parameters pre-optimized. CNN then uses the FCL and soft-max classifiers to extract image features and classify them into the appropriate class. Fig. 14 shows some examples of test images.

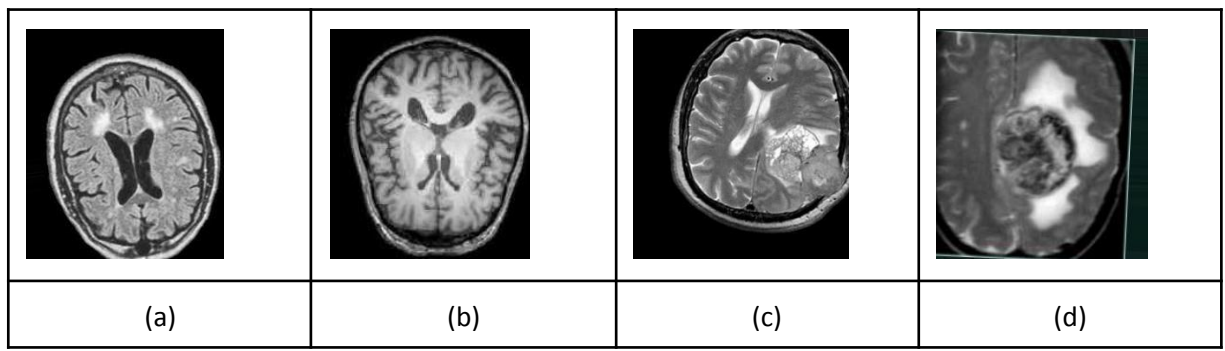

**Fig. 14.** Examples of testing images (a), (b) abnormal images (included tumors), (c), (d) normal images

*6.2. Indicators of performance and assessment criteria*

The performance of the proposed technique in detecting images as normal or abnormal is described in this section. The proposed technique is assessed employing performance indicators such as accuracy, sensitivity, specificity, precision, and F1-score, and receiver operating characteristics (ROC) are utilized to confirm the findings of real-world instances. The performance criteria and mathematical equations are summarized in Table 8. Whereas TP denotes true positivity, TN shows true negativity, FP indicates a false positive, and FN defines a false negative. The accuracy of a classifier is measured by its ability to distinguish between normal and abnormal items in images. TP is the point at which the model anticipates a sufficiently positive specific instance. Therefore, a TN occurs when the model anticipates a negative instance positively. The model incorrectly anticipates a positive position in FP and a negative position in FN. Sensitivity is a measure of the percentage of accurately classified cases. The specificity of a negative occurrence is a measure of how many negative cases are accurately classified. Precision is a metric that measures the proportion of relevant cases among the retrieved samples. It is also known as "positive prediction value." The accuracy of a test is measured by the F1-score, which is characterized as the weighted harmonic mean of its precision and recall.

**Table 8** Performance measures for brain MRI images classification

| Parameters | Formula |
|---|---|
| Accuracy | $\frac{TP+TN}{TP+TN+FP+FN}$ |
| Sensitivity or recall | $\frac{TP}{TP+FN}$ |
| Specificity | $\frac{TN}{TN+FP}$ |
| Precision | $\frac{TP}{TP+FP}$ |
| True Positive Rate | $\frac{TP}{TP+FN}$ |
| False Positive Rate | $\frac{FP}{FP+TN}$ |

| F1 Score | $2\frac{Precision \times Recall}{Precision+Recall}$ |
|---|---|

The receiver operating characteristic (ROC) is a graphical representation of the classification technique's efficiency, taking into account all input factors. This graph demonstrates the connection between 1-specificity and sensitivity.

*6.3. Results and Discussion*

MRI brain images were used to classify the images in this study. MRI images provide helpful information on a variety of tissue parameters (proton density (PD), spin rotation time (T1) and spin rotation time (T2), flow velocity, and chemical change) that lead to more accurate tissue identification. These distinct advantages distinguish MRI as the preferred method for studying brain tumors over other methods. MRI is frequently used for the medical imaging procedure of choice if soft tissue imaging is required. This is particularly important when attempting to classify brain tissues. It is frequently used by radiologists to visualize the internal structure of the body. MRI provides critical information about the anatomy of human soft tissue and aids in diagnosing brain tumors. A "pathological" T2 scan can help identify the affected area of the brain. T1 "anatomical" scans have the highest resolution and are helpful in locating anatomical structures. T2 images were used to diagnose the tumour in this study.

*6.3.1. The confusion matrix and the Receiver operating characteristics (ROC)*

The proposed method for classification effect analysis generates the ROC curve (Fig. 15) and the confusion matrix (Fig. 16). Using the proposed method, 96.0 % of 694 MRI images of the brain were classified as normal and 98.6 % as abnormal (tumor-like). For both cases, the proposed method is highly accurate. As a result, this network can be used to automatically screen brain images. By removing the burden of responsibility in the medical care system, this proposed method significantly aids radiologists in making more accurate diagnoses.

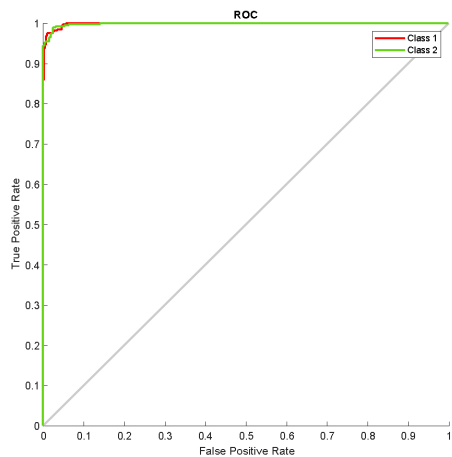

**Fig.15.** NLCMFO-optimized cannulation neural network ROC curve (Class 1 normal, Class 2 abnormal (with tumor))

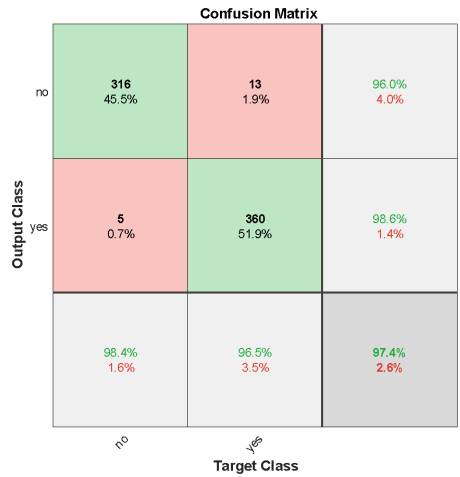

**Fig.16.** NLCMFO optimized convolution neural network confusion matrix

*6.3.2. Progress of the proposed convolutional network training*

Because the selection of hyper-parameters is critical in CNN training, this section compares the results of training operations using the optimized convolutional network with the proposed and non-optimized methods. Fig. 17 depicts the progress of optimized and non-optimized CNN training. In all iterations, the proposed CNN achieves the highest accuracy and the lowest error. Table 9 illustrates the NLCMFO configuration used to optimise the proposed CNN training parameters.

**Table 9** NLCMFO configuration

| Batch Size | Population | Dimensions | Max_Iteration | Lower bound | Upper bound | Hyper parameter cost function |
|---|---|---|---|---|---|---|
| 32 | 30 | 4 | 20 | [0.5, 0.01, 5, 1.0000e-04] | [1, 0.5, 15, 5.0000e-04] | Error rate = (FP + FN)/ (TP + TN + FP + FN) |

*6.4. Performance analysis*

Five standard optimization algorithms besides our proposed NLCMFO have been implemented and their performance was captured and evaluated. PSO (Eberhart & Kennedy, 1995), WOA (Mirjalili & Lewis, 2016),MFO (Mirjalili, 2015), GWO (Mirjalili et al., 2014), and Arithmetic Optimization Algorithm (AOA) (Abualigah et al., 2021) are examples of standard optimization methods. Deep neural network training phase requires 70% of the available data. The rest of the data is consumed during the testing phase. Table 10 compares the accuracy, specificity, sensitivity, Precision, and F1-score of all these networks.

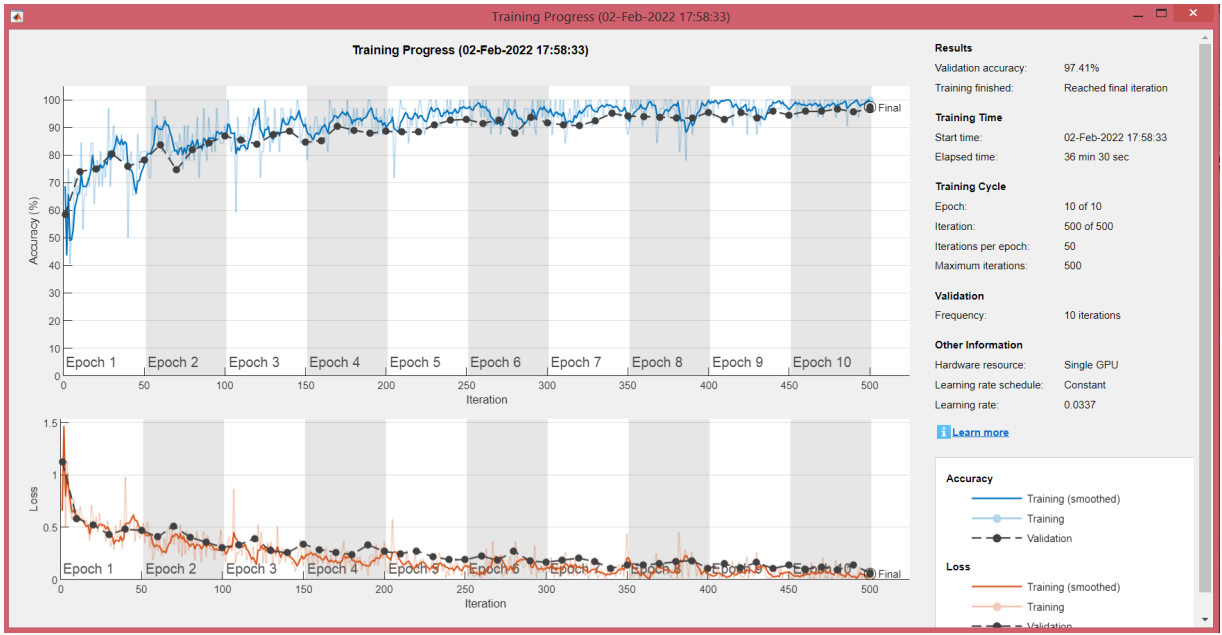

(a)

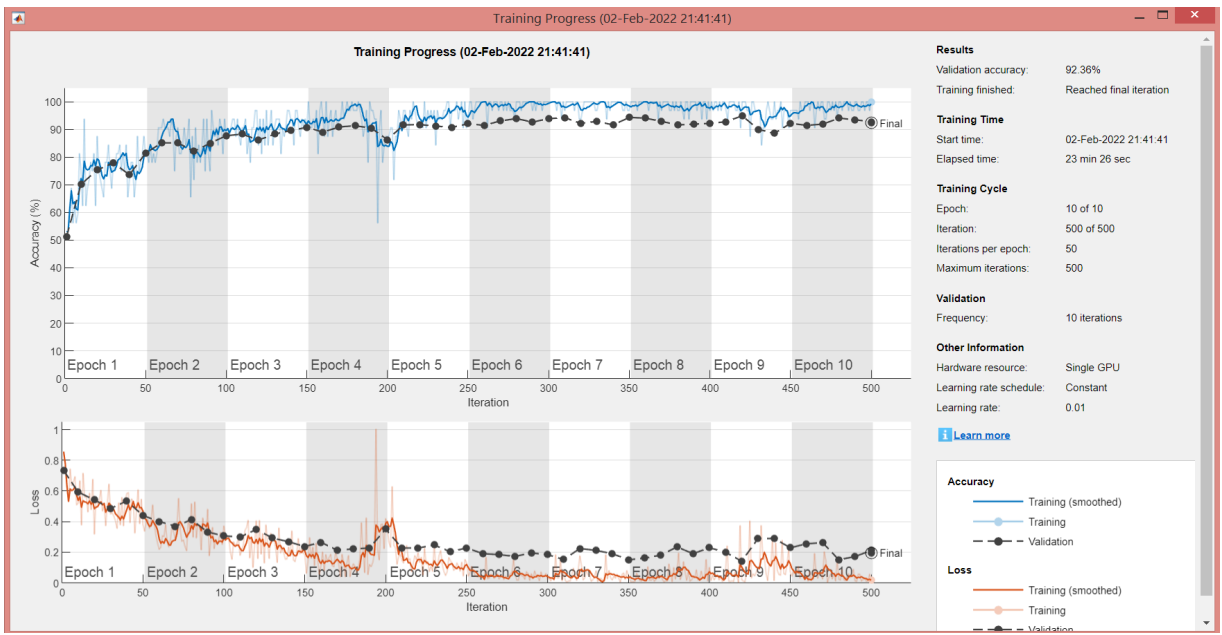

(b)

**Fig.17.** (a) Training progress for NLCMFO-optimized convolutional neural network, (b) Non-optimal convolutional neural network

Figs. 16, 15 show the ROC curve and confusion matrix comparisons. Based on the results of these experiments, the NLCMFO-CNN outperforms all other techniques in terms of accuracy. Therefore, the proposed technique can be employed in medical systems to confidently diagnose brain tumours.

**Table 10** Performance of the proposed method in comparison to existing optimization techniques

| Methods | Accuracy | Sensitivity | Specificity | Precision | F1-score |
|---|---|---|---|---|---|
| Nonoptimization | 92.4% | 89.2% | 95.5% | 95.0% | 90.1% |
| PSO | 96.3% | 95.4% | 97% | 96.6% | 95.99% |
| WOA | 95.8% | 94.2% | 97.3% | 96.9% | 95.53% |
| GWO | 94.1% | 94.9% | 93.5% | 92.2% | 93.14% |
| AOA | 93.8% | 93.7% | 93.9% | 92.8% | 93.24% |
| MFO | 93.2% | 89.4% | 97.1% | 96.9% | 92.9% |
| **NLCMFO** | **97.4%** | **96.0%** | **98.6%** | **98.4%** | **97.18%** |

The results in Table 10 demonstrate that the convolutional neural network optimized by the NLCMFO algorithm outperforms all other methods. As a result, even when solving real-world problems, the proposed method in this study still performs well. The use of optimal hyper-parameters to improve performance and reduce classification error in brain MRI images has been proven in this study, according to the results obtained. Figs. 18 and 19 depict assessment and comparative analysis in both ROC curves and confusion matrices.

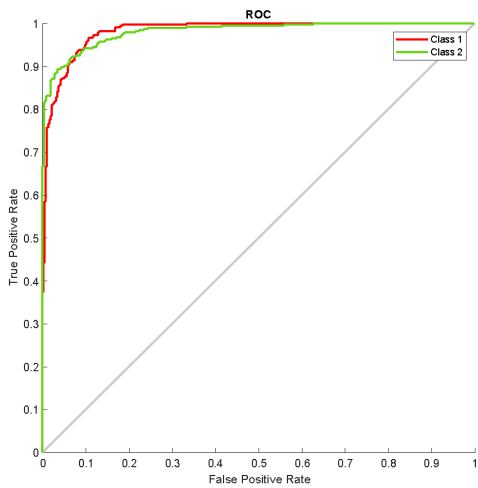
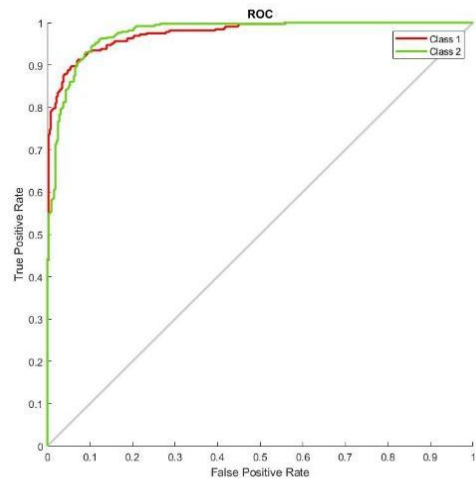

(a)          (b)

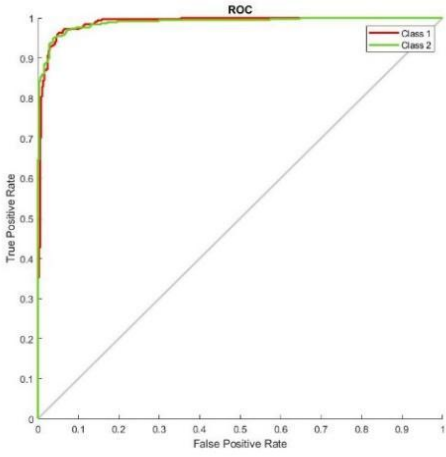 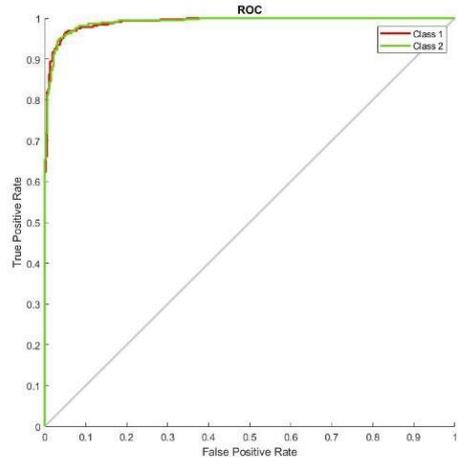

(c)  (d)

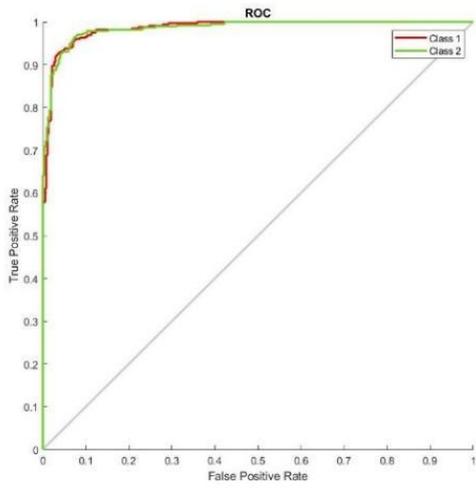 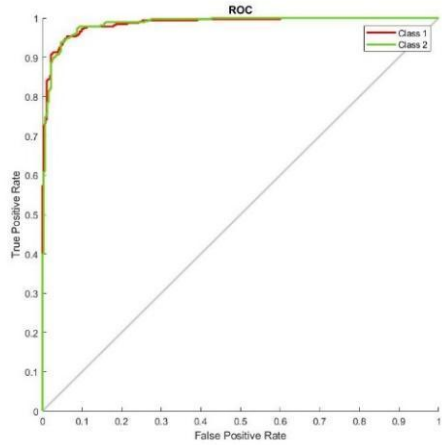

(e)  (f)

**Fig.18.** ROC diagram, (a) non-optimal network, (b) MFO, (c) PSO, (d) WOA, (e) GWO, (f) AOA

(a)

(b)

(c)

(d)

(e)

(f)

**Fig.19.** confusion matrix, (a) non-optimal network, (b) MFO, (c) PSO, (d) WOA, (e) GWO, (f) AOA

The NLCMFO algorithm, used for the automatic classification of brain images in brain MRI images, is used to adjust and optimize the network hyper-parameters in this study. Using the magnitude of the chaotic maps as well as Levy flight theory, the exploration and exploitation phases have been improved. Furthermore, it was able to achieve a proper balance between these two phases by defining a nonlinear weight coefficient parameter. The NLCMFO outperforms the WOA, GWO, PSO, SSA, MVO, and MFO algorithms in the face of 29 standard functions. The results provided in this part indicate that the NLCMFO outperforms the standard MFO and that the hypothesis of using chaotic values, Levy flight theory, and nonlinear weight coefficient parameters to improve its efficiency is valid. The performance of the CNN network optimized by the NLCMFO algorithm has a distinct advantage over other optimization methods, according to the results reported in measuring its performance. The results reveal that the suggested network functioned effectively even when only a few training parameters were used. In terms of accuracy, sensitivity, specificity, precision, and F1-score, the suggested network outperformed the current advanced approaches, confirming the validity of the hypothesis provided in this study.

*6.5. Results analysis*

In a brief overview, the results demonstrate that the standard version of the MFO fared poorly in both challenging test issues and real-world problems involving a multitude of challenges. These observations are also evident in other meta-heuristic algorithms, which explains why such techniques are sufficiently broadly built to tackle various kinds of optimization issues. Therefore, they are applicable to a variety of issues. However, in order to solve a specific set of problems, these algorithms must be modified. The results reported in this section demonstrate that the performance of standard MFO and a number of current optimization approaches is unsatisfactory for addressing significant issues with such a high proportion of dynamic parameters.

The primary goal of this study was to optimise the hyper-parameters of convolutional neural networks, which necessitated a modification of the standard MFO algorithm. This study used optimization issues with a larger set of parameters to assess the MFO's performance, and it has been discovered that the efficiency of this algorithm has significantly decreased in reasonable proportion to the size of the issue. That is partly related to the explosive increase in the solution space in large-scale optimization issues, which require substantial exploration in a meta-heuristic algorithm.There are several approaches throughout the literature for boosting the exploratory behaviour of these algorithms, but the bulk of them also raise the expense of the algorithm's calculating time. As a result, they are less applicable to costly computational optimization problems, such as the ones investigated in this study. Therefore, one of the easiest strategies for improving exploration was chaotic maps. Improving an algorithm's exploration, on the other hand, causes a disruption in its accuracy. This is due to the fact that whenever an algorithm searches a solution space, it finds new areas even without relying on a local search mechanism.

According to preliminary research and results, this problem is caused solely by the use of chaotic maps in MFOs. As a result, we considered using a nonlinear weight coefficient variable to help the MFO improve its accuracy. Furthermore, Levy flight theory was applied to improve the MFO exploitation phase in this study. The benchmark function results demonstrated that the use of Levy flight theory, chaotic maps, and nonlinear mechanisms can significantly improve MFO performance. The proposed NLCMFO outperformed the MFO in the majority of benchmark functions. The superiority of NLCMFO stems from its acceptable exploration and exploitation phases, which employ Levy flight theory, chaotic maps, and a smooth transition to exploitation via a nonlinear mechanism. The results of utilising NLCMFO to tune the hyper-parameters of convolutional neural networks in the classification of brain images into healthy and tumor-bearing demonstrated that NLCMFO can solve this problem with more reliability and less training time than MFO and other algorithms. It should be mentioned that, despite the fact that the vast majority of MFO operators are stored in NLCMFO, the NLCMFO is nevertheless commonly employed to handle a wide range of problems in general. The capacity of NLCMFO to solve optimization problems more successfully is what sets it apart. It is critical to recognise that the proposed method has limitations. To begin with, the NLCMFO algorithm requires the employment of an unique mechanism to cope with restrictions in relatively limited issues. Second, determining the ideal control parameters for the proposed algorithm may necessitate the use of a meta-heuristic algorithm.

## 6. Conclusion

In this study, a chaotic and nonlinear version of the MFO algorithm was proposed in order to optimise the convolutional neural network's hyper-parameters. Despite being a meta-heuristic optimization algorithm, a thorough examination of the MFO algorithm's theoretical foundations revealed its inefficiency when applied to large-scale problems such as optimising the hyper-parameters of convolutional neural networks. Several preliminary experiments were conducted to establish the cause of the MFO's lower accuracy during addressing large-scale issues. The MFO's key weaknesses for such issues were determined to be its exploration and exploitation phases.

Following problem definition, variables such as Levy flight theory and chaotic values were added within the MFO to determine the optimum strategy to maximise the efficiency of the exploration and exploitation processes. However, it was determined that while using Levy flight theory values and chaotic maps improves the algorithm's performance, it is insufficient. To overcome this issue, a nonlinear weight coefficient parameter was devised and incorporated into this algorithm. The NLCMFO performed exceedingly well when compared to a variety of strategies from the literature on a range of benchmark functions. An advantage of this strategy is the employment of nonlinear weight coefficient parameters to achieve the proper equilibrium between both the exploration and exploitation stages. Furthermore, it improves the MFO's exploration and exploitation phases by employing Levy flight theory values and replacing random values with chaotic ones. The NLCMFO was tested on a wide range of uni-modal, multi-modal, and composite benchmark functions employing a wide range of variables. In most experimental cases, the results showed that NLCMFO outperformed MFO and other comparative algorithms. The employment of the nonlinear weight coefficient parameter produced the best possible equilibrium of exploration and exploitation.

Finally, the proposed technique was implemented and its efficiency was assessed in order to identify optimal solutions to the challenging convolutional neural network issue of distinguishing brain MRI images as normal or abnormal (tumor-bearing). NLCMFO has shown promising results in optimising the hyper-parameters of this type of network for brain tumor classification. In terms of results, the NLCMFO significantly outperformed the advanced algorithms, indicating its strengths and benefits. A larger data set is being created as the incidence of brain tumours rises and brain diseases spread around the world. Future research will concentrate on the development and refinement of the proposed network architecture, as well as the combination of newly available data sets for network testing. Furthermore, the proposed algorithm's optimal architecture for the number of layers used in convolutional neural networks can be investigated. Future research could take into account the proposed use of CNN in the classification of brain tumours as furture steps for this study.

APPENDIX A

All benchmark issues utilized in this paper are detailed in Tables 11–14.

**Table 11** Characteristics of the Unimodal function (Rashedi et al., 2009; Yao et al., 1999).

| Func. | Dim | Search Range | $f_{min}$ |
|---|---|---|---|
| $f_1(x) = \sum_{i=1}^{n} x_i^2$ | 30 | [-100,100] | 0 |
| $f_2(x) = \sum_{i=1}^{n} |x_i| + \prod_{i=1}^{n} |x_i|$ | 30 | [-10,10] | 0 |
| $f_3(x) = \sum_{i=1}^{n} \left( \sum_{j-1}^{i} x_j \right)^2$ | 30 | [-100,100] | 0 |
| $f_4(x) = max_i\{|x_i|, 1 \le i \le n\}$ | 30 | [-100,100] | 0 |
| $f_5(x) = \sum_{i=1}^{n-1} \left[ 100(x_{i+1} - x_i^2)^2 + (x_i - 1)^2 \right]$ | 30 | [-30,30] | 0 |
| $f_6(x) = \sum_{i=1}^{n} ([x_i + 0.5])^2$ | 30 | [-100,100] | 0 |
| $f_7(x) = \sum_{i=1}^{n} ix_i^4 + random[0, 1)$ | 30 | [-1.28,1.28] | 0 |

**Table 12** Characteristics of the Multimodal function (Rashedi et al., 2009; Yao et al., 1999).

| Func. | Dim | Search Range | $f_{min}$ |
|---|---|---|---|
| $f_8(x) = \sum_{i=1}^{n} - x_i sin\left(\sqrt{|x_i|}\right)$ | 30 | [-500,500] | -418.9829×n |
| $f_9(x) = \sum_{i=1}^{n} \left[ x_i^2 - 10cos(2\pi x_i) + 10 \right]$ | 30 | [-5.12,5.12] | 0 |
| $f_{10}(x) = -20exp\left(-0.2\sqrt{\frac{1}{n}\sum_{i=1}^{n} x_i^2}\right) - exp\left(\frac{1}{n}\sum_{i=1}^{n} cos(2\pi x_i)\right) + 20 + e$ | 30 | [-32,32] | 0 |
| $f_{11}(x) = \frac{1}{4000}\sum_{i=1}^{n} x_i^2 - \prod_{i=1}^{n} cos\left(\frac{x_i}{\sqrt{i}}\right) + 1$ | 30 | [-600,600] | 0 |
| $f_{12}(x) = \frac{\pi}{n}\left\{10sin(\pi y_1) + \sum_{i}^{n-1} (y_i - 1)^2 \left[ 1 + 10sin^2(\pi y_{i+1}) \right] + (y_n - 1)^2 \right\}$ $y_i = 1 + \frac{x_i+1}{4} u(x_i, a, k, m) = \{k(x_i - a)^m \quad x_i > a \quad 0 - a \quad - a$ | 30 | [-50,50] | 0 |
| $f_{13}(x) = 0.1\left\{ sin^2(3\pi x_i) + \sum_{i=1}^{n} (x_i) - 1^2 \left[ 1 + sin^2(3\pi x_i + 1) \right] + (x_n - 1)^2 \right\}$ | 30 | [-50,50] | 0 |

**Table 13** Characteristics of the Fixed-dimension multimodal function (Rashedi et al., 2009; Yao et al., 1999).

| Func. | Dim | Search Range | $f_{min}$ |
|---|---|---|---|
| $f_{14}(x) = \left( \frac{1}{500} + \sum_{j=1}^{25} \frac{1}{j + \sum_{i=1}^{2}(x_i - a_{ij})^6} \right)^{-1}$ | 2 | [-65,65] | 1 |
| $f_{15}(x) = \sum_{i=1}^{11} \left[ a_i - \frac{x_1(b_i^2 + b_i x_2)}{b_i^2 + b_i x_3 + x_4} \right]^2$ | 4 | [-5,5] | 0.00030 |
| $f_{16}(x) = 4x_1^2 - 2.1x_1^4 + \frac{1}{3}x_1^6 + x_1 x_2 - 4x_2^2 + 4x_2^4$ | 2 | [-5,5] | -1.0316 |
| $f_{17}(x) = \left( x_2 - \frac{5.1}{4\pi^2}x_1^2 + \frac{5}{\pi}x_1 - 6 \right)^2 + 10\left(1 - \frac{1}{8\pi}\right)\cos x_1 + 10$ | 2 | [-5,5] | 0.398 |
| $f_{18}(x) = \left[ 1 + (x_1 + x_2 + 1)^2 (19 - 14x_1 + 3x_1^2 - 14x_2 + 6x_1 x_2 + 3x_2^2) \right] \times [30 \cdots]$ | 2 | [-2,2] | 3 |
| $f_{19}(x) = -\sum_{i=1}^{4} c_i \exp\left( -\sum_{j=1}^{3} a_{ij}(x_j - p_{ij})^2 \right)$ | 3 | [1,3] | -3.86 |
| $f_{20}(x) = -\sum_{i=1}^{4} c_i \exp\left( -\sum_{j=1}^{6} a_{ij}(x_j - p_{ij})^2 \right)$ | 6 | [0,1] | -3.32 |
| $f_{21}(x) = -\sum_{i=1}^{5} \left[ (X - a_i)(X - a_i)^T + c_i \right]^{-1}$ | 4 | [0,10] | -10.1532 |
| $f_{22}(x) = -\sum_{i=1}^{7} \left[ (X - a_i)(X - a_i)^T + c_i \right]^{-1}$ | 4 | [0,10] | -10.4028 |
| $f_{23}(x) = -\sum_{i=1}^{10} \left[ (X - a_i)(X - a_i)^T + c_i \right]^{-1}$ | 4 | [0,10] | -10.5363 |

**Table 14** Characteristics and an overview of the composition functions of the CEC-BC-2014 (Please see (Liang et al., 2014) for further information).

| | No. | Func. | Dim | Search Range | $f_{min}$ |
|---|---|---|---|---|---|
| Composition Func. | 1 | Composition Func. (24) | 30 | [-100,100] | 2400 |
| | 2 | Composition Func. (25) | 30 | [-100,100] | 2500 |
| | 3 | Composition Func. (26) | 30 | [-100,100] | 2600 |
| | 4 | Composition Func. (27) | 30 | [-100,100] | 2700 |
| | 5 | Composition Func. (28) | 30 | [-100,100] | 2800 |
| | 6 | Composition Func. (29) | 30 | [-100,100] | 2900 |